\documentclass[fleqn,10pt]{wlscirep}
\usepackage[utf8]{inputenc}
\usepackage[T1]{fontenc}
\usepackage{amsmath,graphicx,bm}
\usepackage{amssymb}
\usepackage[utf8]{inputenc}
\usepackage{newunicodechar}
\usepackage{multirow}
\usepackage{subcaption}
\usepackage{amsfonts}
\usepackage{bbm}
\usepackage{makecell}
\usepackage{balance}
\usepackage{lineno}
\usepackage[inkscapeformat=png]{svg}
\usepackage{amssymb}
\usepackage{latexsym}
\usepackage{url}
\usepackage{xcolor}
\usepackage{hyperref}

\title{Spatio-Temporal Graph Deep Learning with Stochastic Differential Equations for Uncovering Alzheimer's Disease Progression}

\author[1]{Houliang Zhou}
\author[1]{Rong Zhou}
\author[1]{Yangying Liu}
\author[2]{Kanhao Zhao}
\author[3]{Li Shen}
\author[1]{Brian Y. Chen}
\author[4,*]{Yu Zhang}
\author[1,*]{Lifang He}
\author[5]{Alzheimer's Disease Neuroimaging Initiative}
\affil[1]{Department of Computer Science and Engineering, Lehigh University, PA, USA}
\affil[2]{Department of Bioengineering, Lehigh University, PA, USA}
\affil[3]{Department of Biostatistics, Epidemiology and Informatics, University of Pennsylvania, PA, USA}
\affil[4]{Department of Psychiatry and Behavioral Sciences, Stanford University School of Medicine, CA, USA}
\affil[5]{Alzheimer's Disease Neuroimaging Initiative}
\affil[*]{Correspondence: yzhangsu@stanford.edu, lih319@lehigh.edu}


\begin{abstract}
Identifying objective neuroimaging biomarkers to forecast Alzheimer's disease (AD) progression is crucial for timely intervention. However, this task remains challenging due to the complex dysfunctions in the spatio-temporal characteristics of underlying brain networks, which are often overlooked by existing methods. To address these limitations, we develop an interpretable spatio-temporal graph neural network framework to predict future AD progression, leveraging dual Stochastic Differential Equations (SDEs) to model the irregularly-sampled longitudinal functional magnetic resonance imaging (fMRI) data. We validate our approach on two independent cohorts, including the Open Access Series of Imaging Studies (OASIS-3) and the Alzheimer's Disease Neuroimaging Initiative (ADNI). Our framework effectively learns sparse regional and connective importance probabilities, enabling the identification of key brain circuit abnormalities associated with disease progression. Notably, we detect the parahippocampal cortex, prefrontal cortex, and parietal lobule as salient regions, with significant disruptions in the ventral attention, dorsal attention, and default mode networks.  These abnormalities correlate strongly with longitudinal AD-related clinical symptoms. Moreover, our interpretability strategy reveals both established and novel neural systems-level and sex-specific biomarkers, offering new insights into the neurobiological mechanisms underlying  AD progression. Our findings highlight the potential of spatio-temporal graph-based learning for early, individualized prediction of  AD progression, even in the context of irregularly-sampled longitudinal imaging data.

\end{abstract}

\begin{document}
\flushbottom
\maketitle

%
%


\section*{Introduction}
\label{sec1}
Recent advances in neuroimaging have enabled the collection of longitudinal functional magnetic resonance imaging (fMRI) data \cite{skup2010longitudinal}, offering crucial insights into individual trajectories of neurological disorders such as Alzheimer's disease \cite{crone2015changing}. Compared to cross-sectional analyses, longitudinal imaging provides a more accurate characterization of brain development and disease progression. For example, studies have shown that hippocampal volume declines significantly over time in aging populations, a trend that is often overlooked in cross-sectional studies  \cite{pfefferbaum2015cross, di2023mapping}. Longitudinal fMRI captures within-individual changes in functional connectivity, which allows for tracking disease progression and modeling dynamic reorganization of neural systems \cite{lamontagne2019oasis}. By constructing functional brain networks, where nodes represent anatomical regions of interest (ROIs) and edges reflect statistical correlations between regional fMRI time series, longitudinal studies provide a powerful framework for understanding disease-related connectivity abnormalities and modeling disease progression in neurodegenerative conditions, such as AD and mild cognitive impairment (MCI) \cite{meskaldji2016prediction, li2021braingnn}.

Learning spatio-temporal patterns of brain activity is critical for advancing the detection and understanding of neurological disorders.
In the context of AD, these patterns are characterized by the progressive spread of dysfunction across brain regions over time \cite{meyer2022spatiotemporal,kim2024distinct}. As AD progresses to the later stage, distinct trajectories of cortical thinning emerge, particularly in the medial temporal lobe \cite{kim2024distinct}. Such patterns also help delineate the spatial progression of atrophy, functional disconnections, and pathological accumulation such as amyloid-beta and tau across different disease stages \cite{collij2022spatial, hampel2021amyloid}. Deep learning, particularly graph neural networks (GNNs), has demonstrated remarkable capabilities in analyzing spatio-temporal patterns of brain activity to detect neurological disorders \cite{parisot2018disease,cui2021brainnnexplainer,zhou2022interpretable, zhou2023interpretable, zhou2024multi}. GNN-based approaches such as BrainGNN have leveraged ROI-aware graph convolutional layers to identify disease-specific biomarkers from fMRI data \cite{li2021braingnn}, while subgraph-based models have applied information bottleneck principles to improve biomarker detection \cite{zheng2024brainib}. However, these methods primarily analyze static connectivity patterns and fail to capture the temporal evolution of brain networks, limiting their ability to predict AD progression. Given that neurodegenerative diseases like AD manifest through progressive disruptions in brain networks, there is a pressing need for spatio-temporal graph learning methods that can model dynamic brain network changes over time. Furthermore, explainability in GNN models is essential for uncovering interpretable biomarkers, as demonstrated by explainable GNNs that identify disorder-specific features linked to neural circuit dysfunction and cognition \cite{cui2021brainnnexplainer}. 

Several statistical and machine learning approaches have been developed to model disease trajectories. Event-based models, for instance, estimate the sequential progression of biomarkers in AD \cite{venkatraghavan2017discriminative}, while deep learning frameworks have been designed to predict clinical symptoms from longitudinal imaging data \cite{lei2020deep}. However, these methods often treat imaging data as a time series of hand-crafted features, ignoring the irregular time intervals between scans and the temporal sparsity of images for individual patients. In clinical settings, patients undergo imaging at variable time points, making it challenging to learn continuous disease dynamics. Recently, the stochastic differential equation (SDE), as a generalized form of an ordinary differential equation (ODE) by incorporating random perturbations, emerged as promising tools for modeling continuous data and dynamic systems in applications to physical system simulation and disease spread modeling \cite{huang2020learning, huang2023generalizing}. For example, ImageFlowNet employs neural ODEs and SDEs to model disease progression in the retinal image and brain multiple sclerosis MRI scans \cite{liu2024imageflownet}, but it ignores the spatial information of dynamic brain networks, which is crucial for understanding network-level abnormalities. Similarly, BrainODE effectively reconstructs brain signals at arbitrary time points \cite{han2024brainode}, yet it did not operate in the longitudinal domain, limiting its ability to identify potential biomarkers for tracking disease progression trajectories. Most existing approaches analyze spatial and temporal aspects independently \cite{kiviniemi2011sliding, liu2024imageflownet}, despite the fact that these dimensions are deeply interconnected in neurodegenerative progression \cite{zhang2021spatiotemporal}. A unified spatio-temporal framework is essential for capturing the intricate brain dysfunctions underlying AD.

\begin{figure*}[!ht]
\centering
\includegraphics[width=\linewidth]{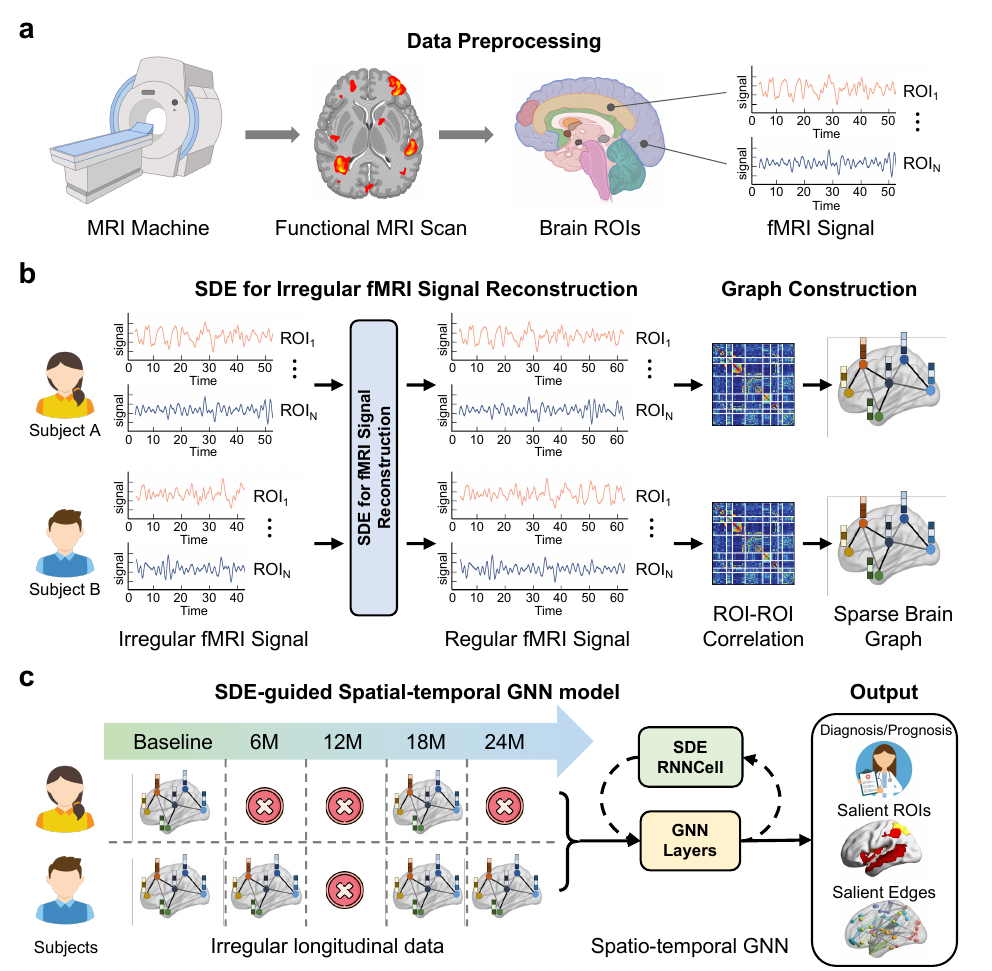}
\caption{An overview of our proposed model for AD progression prediction and biomarker interpretation. {\bf (a)} The datasets consist of longitudinal fMRI scans, with up to six time points spanning from baseline to 105 months of follow-up. {\bf (b)} The longitudinal fMRI scans are preprocessed and reconstructed to continuous signals for each ROI by the SDE method. The dynamic graphs were built by using longitudinal reconstructed fMRI signals. First, the connections between ROIs were quantified by measuring the correlation between their reconstructed signals. Next, the reconstructed signals in the ROIs, along with their connections, were combined with the proposed learnable importance probabilities to generate a sparse graph. The dynamic graphs thus consist of longitudinal sparse graphs. {\bf (c)} The dynamic graphs are sent to our SDE-guided spatio-temporal GNN model to learn and evolve the longitudinal representation. The learned representation is fed into an MLP classifier to predict the disease progression. The importance probabilities on nodes and edges of sparse graphs provide the interpretation for the salient ROIs and the prominent disease-specific connections.}
\label{fig:architecture_overview}
\end{figure*}

To address these challenges, we propose an interpretable SDE-guided spatio-temporal GNN framework (SDE-GNN) for predicting AD progression and uncovering underlying neurobiological mechanisms. Our model learns longitudinal functional connectivity patterns from fMRI, allowing for continuous inference across irregularly sampled imaging data. This innovative approach integrates temporal evolution modeling with graph-based learning to improve predictive accuracy and interpretability. Our framework demonstrates high accuracy in detecting progressive AD patients using data from the large-scale Open Access Series of Imaging Studies (OASIS-3) cohort \cite{marcus2010open}. Using our interpretation technique, we pinpoint the parahippocampal cortex, prefrontal cortex, and parietal lobule as key regions implicated in the early stage of AD progression. Critical abnormal connections primarily involve ventral attention, dorsal attention, and default mode networks showing statistical significance in distinguishing progressive patients from stable ones. These longitudinal connectivity biomarkers exhibit strong correlations with the trajectories of cognitive decline and amyloid-beta. Sex-related differences in these connectivity biomarkers are also examined. Importantly, our framework and biomarker findings are replicated using independent data from the Alzheimer’s Disease Neuroimaging Initiative (ADNI) \cite{mueller2005alzheimer}. Together, our study provides a powerful tool for predicting AD progression from irregularly sampled fMRI data and offers new insights into disease-specific brain network alterations. By integrating spatiotemporal graph learning with explainable deep learning techniques, our framework paves the way for early diagnosis, personalized intervention, and biomarker discovery in neurodegenerative research.


\begin{figure*}[!t]
\centering
\includegraphics[width=\linewidth]{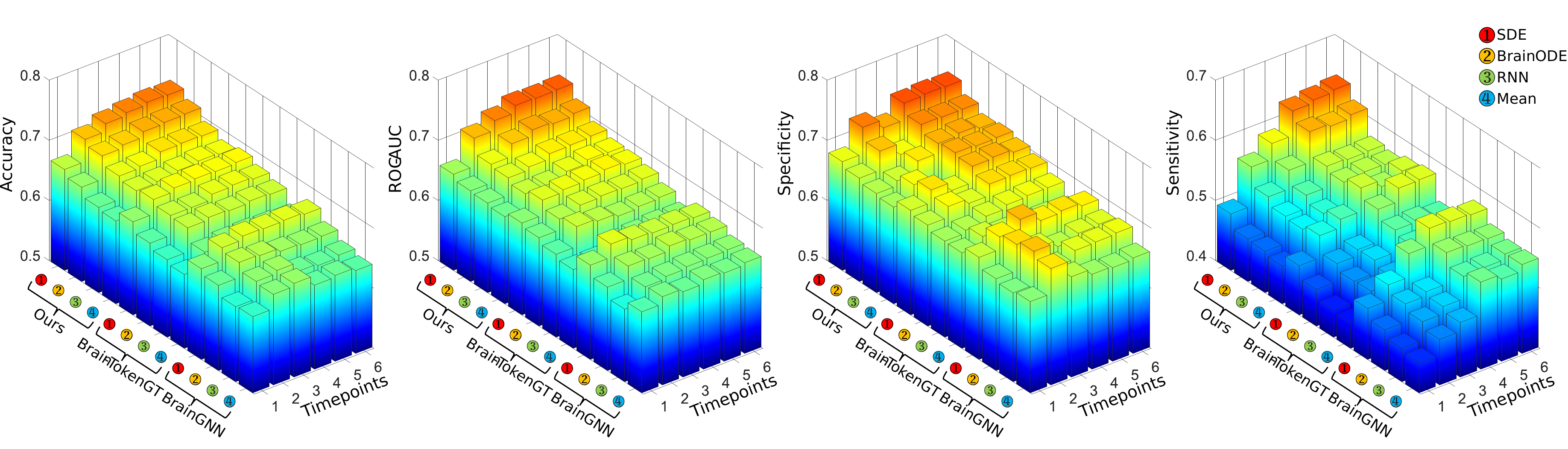}
\caption{Classification comparison of stable vs. progressive subjects in OASIS-3 between the state-of-the-art machine learning models and ours by using different preprocessing strategies including SDE, BrainODE, RNN, and Mean. The average classification accuracy, the area under the receiver operating characteristic curve (ROC-AUC), specificity, and sensitivity are reported under the 5-fold cross validation. The best performance was achieved by using our proposed method.}
\label{fig:classification_oasis}
\end{figure*}

\section*{Results}

\subsection*{Longitudinal connectome-based predictive modeling of AD progression}

We used longitudinal fMRI data from the Open Access Series of Imaging Studies (OASIS-3) \cite{marcus2010open} as a discovery sample to evaluate the efficacy of our method for predicting AD progression . The dataset included 554 stable subjects and 193 progressive subjects, each with up to six time points spanning from baseline to 105 months of follow-up. Our prediction experiment aims to distinguish stable individuals from those who would later progress to MCI or AD, using only pre-progression time points. The progressive group comprised subjects who eventually converted from healthy controls (HC) to MCI or AD, and those from MCI to AD. This experimental design enables an evaluation of our approach's ability to predict future disease progression. An overview of our approach for predicting Alzheimer's progression and interpreting potential biomarkers is shown in Fig. \ref{fig:architecture_overview}. Our method first employed a neural SDE approach to reconstruct missing values in fMRI signals (Fig. \ref{fig:architecture_overview}b). We then computed correlations between reconstructed regional fMRI time series to create brain connectivity graphs at each time point. Given the irregularly-sampled longitudinal data, we integrated these time-specific graphs to form dynamic brain network representations. Our spatio-temporal GNN model learned longitudinal connectivity patterns from dynamic graphs to predict disease progression (Fig. \ref{fig:architecture_overview}c). Meanwhile, we applied an importance probability approach to identify key ROIs and connections associated with disease progression. For evaluation, we performed a 5-fold cross-validation and assessed performance using multiple classification metrics, including average accuracy, area under the receiver operating characteristic curve (ROC-AUC), sensitivity, and specificity.


Fig. \ref{fig:classification_oasis} presents a comparative analysis of classification performance between our proposed model and the state-of-the-art machine learning approaches for predicting AD progression. We systematically evaluated the impact of different timeseries reconstruction strategies for building missing fMRI signals and compared various longitudinal connectome-based learning methods. In the preprocessing step, we compared neural SDE approach against three alternative methods: (a) Mean imputation, a commonly used baseline method that fills missing values with the average of the observed signals. (b) Recurrent Neural Network (RNN), which models sequential dependencies by maintaining a hidden state to capture temporal patterns \cite{sherstinsky2020fundamentals}. (c) BrainODE, a state-of-the-art method that employs Ordinary Differential Equations (ODE) to continuously model dynamic brain signals \cite{han2024brainode}. Following the reconstructing of missing fMRI signals, we applied our proposed spatio-temporal GNN model to learn longitudinal connectome-based representations and predict AD progression. For benchmarking, we compared our approach against two leading GNN-based methods for brain network analysis: BrainGNN \cite{li2021braingnn}, and BrainTokenGT \cite{dong2023beyond}. Our result indicated that the SDE-based timeseries reconstruction strategy improved accuracy, sensitivity, and specificity by approximately 3\%, 6\%, and 4\%, respectively, compared to other alternative methods. In longitudinal modeling, spatio-temporal GNN model outperformed both BrainGNN and BrainTokenGT, achieving 3\% to 5\% higher accuracy, sensitivity, and specificity. Notably, the highest classification performance was obtained using our full pipeline, highlighting the effectiveness of integrating SDE-based reconstruction with spatio-temporal graph learning for AD progression prediction, . Building on these classification results, we further applied our method to evaluate interpretability and identify potential neurobiological biomarkers associated with  AD progression.

\begin{figure*}[!t]
\centering
\includegraphics[width=\linewidth]{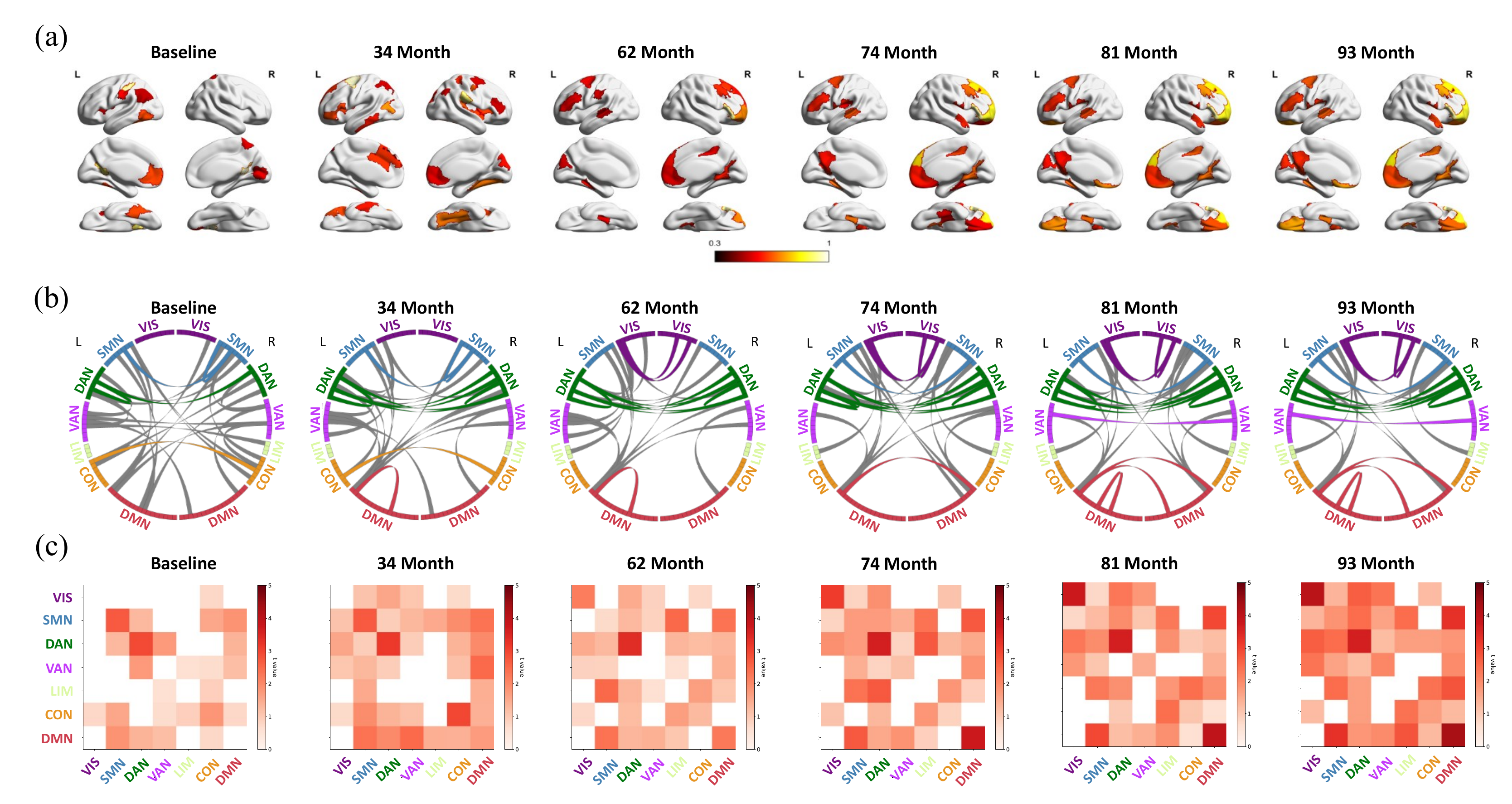}
\caption{Interpretation analysis in OASIS-3 cohort. {\bf (a)} Interpreting top 20 selected salient ROIs in the progressive group across six models, each trained on an increasing number of timesteps. The title indicates the mean follow-up months of subjects. The color bar ranges from 0.3 to 1.0. The bright-yellow color indicates a high score, while dark-red color indicates a low score. {\bf (b)} The significant difference of the interpreted most discriminative connections for distinguishing stable and progressive subjects was evaluated by two-sample t-tests with FDR corrected p-value $<$ 0.05. Here, the top 30 most discriminative ROI connections are visualized for interpretation over 6 different timesteps. The intra-network connections are colored based on the module itself and inter-network connections are colored in grey. {\bf (c)} Neural system-level interpretation of the most discriminative connections was computed by averaging the absolute t value of most discriminative ROI connections reported between neural systems over 6 different timesteps. The dark-red color indicates significant connections. The non-significant connections are marked as white. VIS = Visual Network, SMN = Somatomotor Network, DAN = Dorsal Attention Network, VAN = Ventral Attention Network, LIM = Limbic Network, CON = Control Network, DMN = Default Mode Network.}
\label{fig:oasis_interpretation}
\end{figure*}

\subsection*{Mapping key brain circuit abnormalities underlying AD progression}
To identify key ROIs and connectivity abnormalities associated with AD progression, we  analyzed sparse regional and connective importance probabilities learned by our method. These importance probabilities provide an interpretable framework for detecting disease-related ROIs and connectivity disruptions, offering insights into the neurobiological mechanisms underlying AD progression. We ranked regional importance probabilities in descending order to determine the most salient ROIs for distinguishing stable from progressive subjects. Fig. \ref{fig:oasis_interpretation}(a) presents the top 20 most discriminative ROIs across all longitudinal time points. Our findings highlight the parahippocampal cortex, lateral and medial prefrontal cortex, extrastriate superior cortex, and insula as key regions predictive of AD progression, aligning with prior neuroimaging evidence \cite{dickerson2009differential,mu2011adult,eskildsen2015structural}. The results revealed that different brain regions play distinct roles at different disease stages. Specifically, the auditory cortex and superior parietal lobule exhibited higher importance in the early stage of progression, whereas the lateral and medial prefrontal cortex became more influential at later stages. Notably, the parahippocampal cortex consistently remained a highly salient ROI from baseline through all follow-up time points, suggesting its critical role in AD progression.
 

To identify discriminative functional connections, we analyzed connective importance probabilities and performed two-sample $t$-tests to detect statistically significant differences between stable and progressive groups. Fig. \ref{fig:oasis_interpretation}(b) displays the top 30 most discriminative connections (false discovery rate (FDR) correlated p-value $<$ 0.05) from baseline to all the follow-up months. To facilitate neuroscientific interpretation, we grouped ROIs into seven major neural systems based on the Schaefer100 atlas  \cite{yeo2011organization}, including visual (\textcolor[RGB]{120,18,134}{VIS}), somatomotor (\textcolor[RGB]{70,130,180}{SMN}), dorsal attention  (\textcolor[RGB]{0,118,14}{DAN}), ventral attention  (\textcolor[RGB]{196,58,250}{VAN}), limbic  (\textcolor[RGB]{220,248,164}{LIM}), frontopartiel control  (\textcolor[RGB]{230,148,34}{CON}), and default mode networks (\textcolor[RGB]{205,62,78}{DMN}). In Fig. \ref{fig:oasis_interpretation}(b), intra-network connections are colored according to their respective neural systems, while inter-network connections are colored in grey. Our findings reveal that the most discriminative connections were centered around extrastriate cortex in visual network, auditory in somatomotor network, post central cortex and superior parietal lobule in dorsal attention network, prefrontal cortex and temporal in default mode network from baseline to all the follow-up months. These connectivity alterations are consistent with recent studies demonstrating functional disruptions in the dorsal attention and default mode networks in AD \cite{schwab2020functional}. While no significant disruptions were observed in the visual network at early stages, abnormalities in the extrastriate cortex emerged as AD progressed, indicating a late-stage connectivity breakdown in the visual system.

To further explore disease-related neural system-level disruptions, we examined the inter- and intra-network connectivity patterns most strongly associated with AD progression. Fig. \ref{fig:oasis_interpretation}(c) depicts a heatmap of absolute $t$ values, illustrating differential neural system connections across all time points. Our findings reveal persistent connectivity abnormalities in the dorsal attention and default mode networks across all disease stages, reinforcing their role in AD-related functional deterioration. Interestingly, no connectivity dysfunctions were observed within the default mode and visual networks when using only the baseline data, while some dysfunctions were concentrated within them at the later stages. Meanwhile, most discriminative connections within visual and default mode networks were found when using longitudinal data instead of the baseline. These patterns suggested that functional disruptions in these networks intensify shortly before clinical progression, highlighting their potential as late-stage biomarkers. Moreover, our results show that longitudinal data significantly enhanced the detection of these neural system-level abnormalities, supporting the value of multi-timepoint analysis in AD research. These findings are consistent with previous neuroimaging studies demonstrating progressive connectivity dysfunctions in AD and MCI \cite{schwab2020functional,lerch2005focal}. Our results further demonstrate that integrating irregularly sampled longitudinal fMRI data improves the detection of functional network disruptions, underscoring the power of our model in capturing AD-related brain dynamics.

\subsection*{Replication analysis on an independent cohort}
To replicate our longitudinal predictive modeling and connectome biomarker findings, we conduct a validation study using ADNI cohort, employing the same analytical pipeline as the OASIS-3 discovery cohort. The ADNI cohort included 261 stable and 33 progressive subjects, each with up to six longitudinal time points spanning from baseline to 54 months of follow-up. Using the same methodology as the discovery cohort, we constructed the longitudinal irregularly-sampled brain connectivity graphs for each subject and applied our spatio-temporal GNN model to learn spatial and temporal connectivity patterns from these graphs for distinguishing progressive versus stable subjects. Additionally, we leveraged importance probability analysis to identify key neural circuits associated with disease progression. To evaluate model performance, we performed 5-fold cross-validation and assessed key classification metrics, including accuracy, ROC-AUC, sensitivity, and specificity.

\begin{figure*}[!t]
\centering
\includegraphics[width=\linewidth]{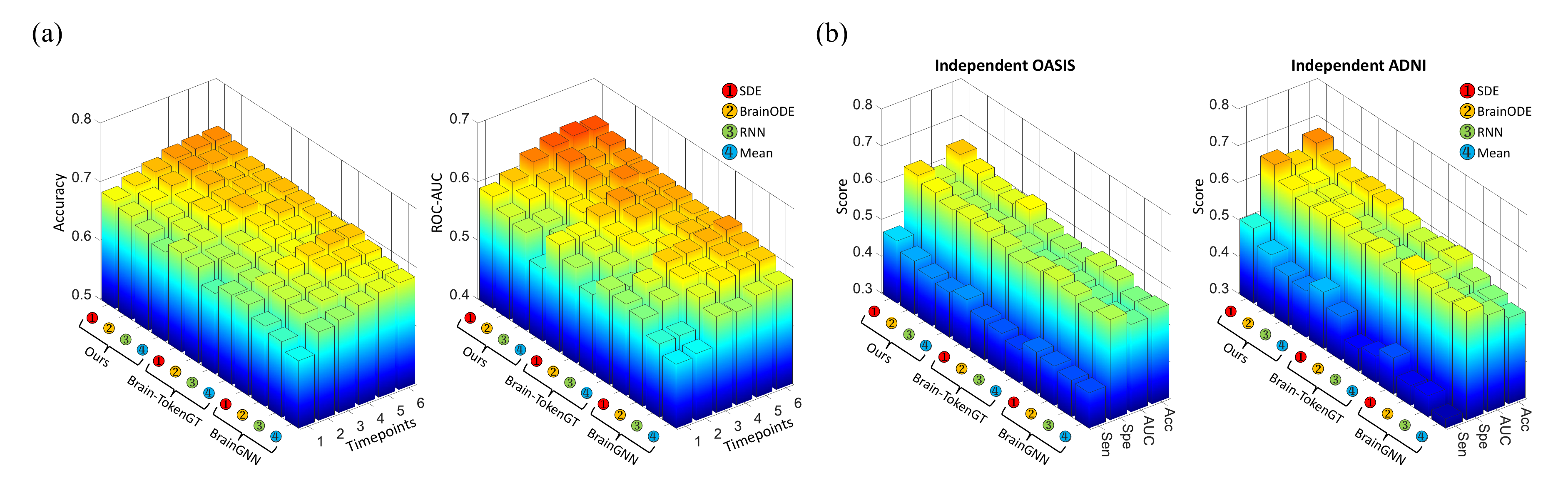}
\caption{{\bf (a)} Classification comparison of stable vs. progressive contrast in ADNI validation cohort between the state-of-the-art machine learning models and ours by using different preprocessing strategies including SDE, BrainODE, RNN, and Mean. {\bf (b)} Classification performance of independent ADNI and OASIS-3 test cohort using all available timepoints. The average classification accuracy (Acc), the area under the receiver operating characteristic curve (ROC-AUC), specificity (Spe), and sensitivity (Sen) are reported under the 5-fold cross-validation.}
\label{fig:classification_adni}
\end{figure*}

Fig. \ref{fig:classification_adni}(a) shows the classification comparison between the state-of-the-art machine learning models and our proposed model in the ADNI cohort. We perform the same analysis and comparisons as in the discovery cohort to evaluate different reconstruction methods and brain-related GNN-based approaches. Consistent with the prediction performance from the discovery data, our results demonstrate that the SDE reconstruction method consistently improves accuracy, sensitivity, and specificity by 2\% to 4\% compared to other reconstruction strategies. Similarly, in longitudinal learning, our model achieves a 2\% to 5\% increase in accuracy and ROC-AUC compared to BrainGNN and BrainTokenGT, reaffirming the advantages observed in the discovery cohort. Notably, the best accuracy is achieved using our method with longitudinal data, and its prediction performance closely matches the results obtained in the OASIS-3 cohort, further validating its robustness and reliability. To further investigate the generalizability of our method, we evaluated its performance on an independent test set. Specifically, we trained our model on the OASIS-3 cohort and tested it independently on the ADNI cohort, and vice versa. Fig. \ref{fig:classification_adni}(b) presents the classification performance when using ADNI and OASIS-3 as independent test sets. Our findings indicate that the SDE preprocessing method achieves an accuracy improvement of over 3\% compared to other preprocessing strategies. Furthermore, our model consistently outperforms other methods in capturing longitudinal information for disease progression prediction. An ablation study was conducted to validate the effectiveness of SDE over latent ODE in Spatial-temporal GNN model, reported in Supplementary Fig. \ref{fig:ablation_study}. Based on these classification results, our method was used to evaluate the interpretability and find the biomarkers in ADNI.

\begin{figure*}[!t]
\centering
\includegraphics[width=\linewidth]{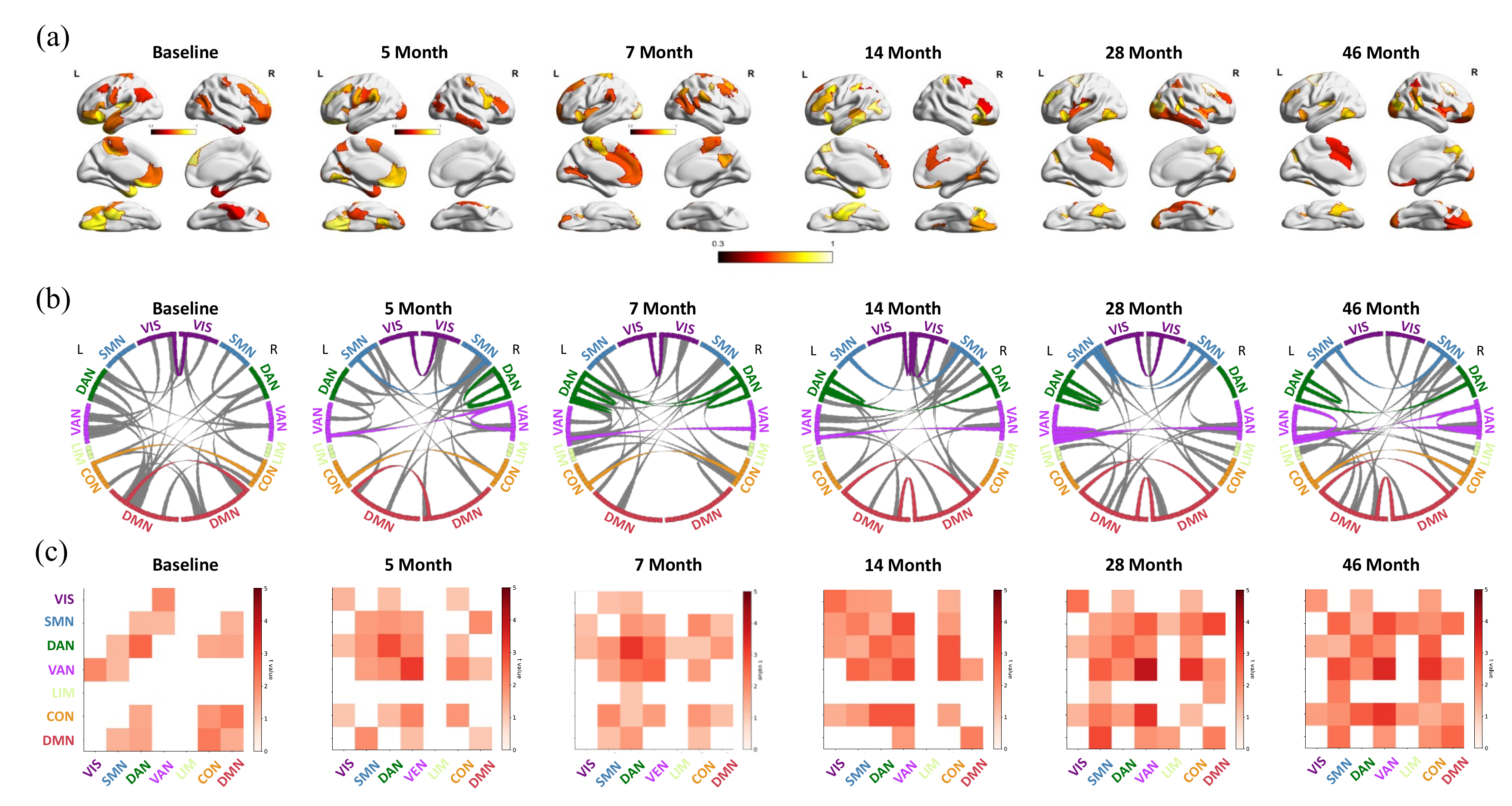}
\caption{Reproducibility of interpretation in validation ADNI cohort. (a) Interpreting top 20 selected salient ROIs in the progressive group across six models, each trained on an increasing number of timesteps. The title indicates the mean follow-up months of subjects. The color bar ranges from 0.3 to 1.0. The bright-yellow color indicates a high score, while dark-red color indicates a low score. (b) The significant difference of the interpreted most discriminative connections for distinguishing stable and progressive subjects was evaluated by two-sample t-tests with FDR corrected p-value $<$ 0.05. Here, the top 30 most discriminative ROI connections are visualized for interpretation over 6 different timesteps. The intra-network connections are colored based on the module itself and inter-network connections are colored in grey. (c) Neural system-level interpretation of the most discriminative connections was computed by averaging the absolute t value of most discriminative ROI connections reported between neural systems over 6 different timesteps. The dark-red color indicates a high score. The non-significant connections are marked as white. VIS = Visual Network, SMN = Somatomotor Network, DAN = Dorsal Attention Network, VAN = Ventral Attention Network, LIM = Limbic Network, CON = Control Network, DMN = Default Mode Network.}
\label{fig:adni_interpretation}
\end{figure*}

We conducted the same interpretability analysis on the ADNI cohort to identify key ROIs and discriminative functional connections, validating our biomarker findings from the OASIS-3 cohort. Fig. \ref{fig:adni_interpretation} presents the interpretation results for the ADNI data. Fig. \ref{fig:adni_interpretation}(a) highlights the parahippocampal cortex, post-central cortex, prefrontal cortex, and parietal lobule as key brain regions associated with AD progression. Fig. \ref{fig:adni_interpretation}(b) illustrates the interpretation of brain connectivity abnormalities, revealing that the most discriminative connections involve the post-central lobule in the dorsal attention network, the temporal-occipital and parietal-frontal medial regions in the ventral attention network, and the temporal and prefrontal cortex in the default mode network. Specifically, the most discriminative connections around the post-central lobule and parietal-frontal medial regions in the dorsal and ventral attention networks become stronger at a later stage, while the connections around the prefrontal cortex and temporal regions in the default mode network remain consistently strong over all stages. Similar findings are observed at the neural system level, as reported in Fig. \ref{fig:adni_interpretation}(c). These results from the ADNI validation cohort reinforce our previous findings from the OASIS-3 discovery cohort, highlighting that both intra and inter neural system-level connections within the attention network and default mode network play a critical role in AD progression.

To further investigate the consistency between biomarkers discovery across cohorts, we assessed the correlation of regional and connective importance probabilities identified in OASIS-3 and ADNI. The correlation coefficient was 0.46 for salient ROIs and 0.55 for discriminated connections, indicating a strong agreement between the two independent datasets. The statistical significance of these correlations was confirmed through 10,000 random permutation tests, ensuring the robustness of our findings. Further details are summarized in Supplementary Fig. \ref{fig:corr_biomarker}.

\begin{figure*}[!ht]
\centering
\includegraphics[width=\linewidth]{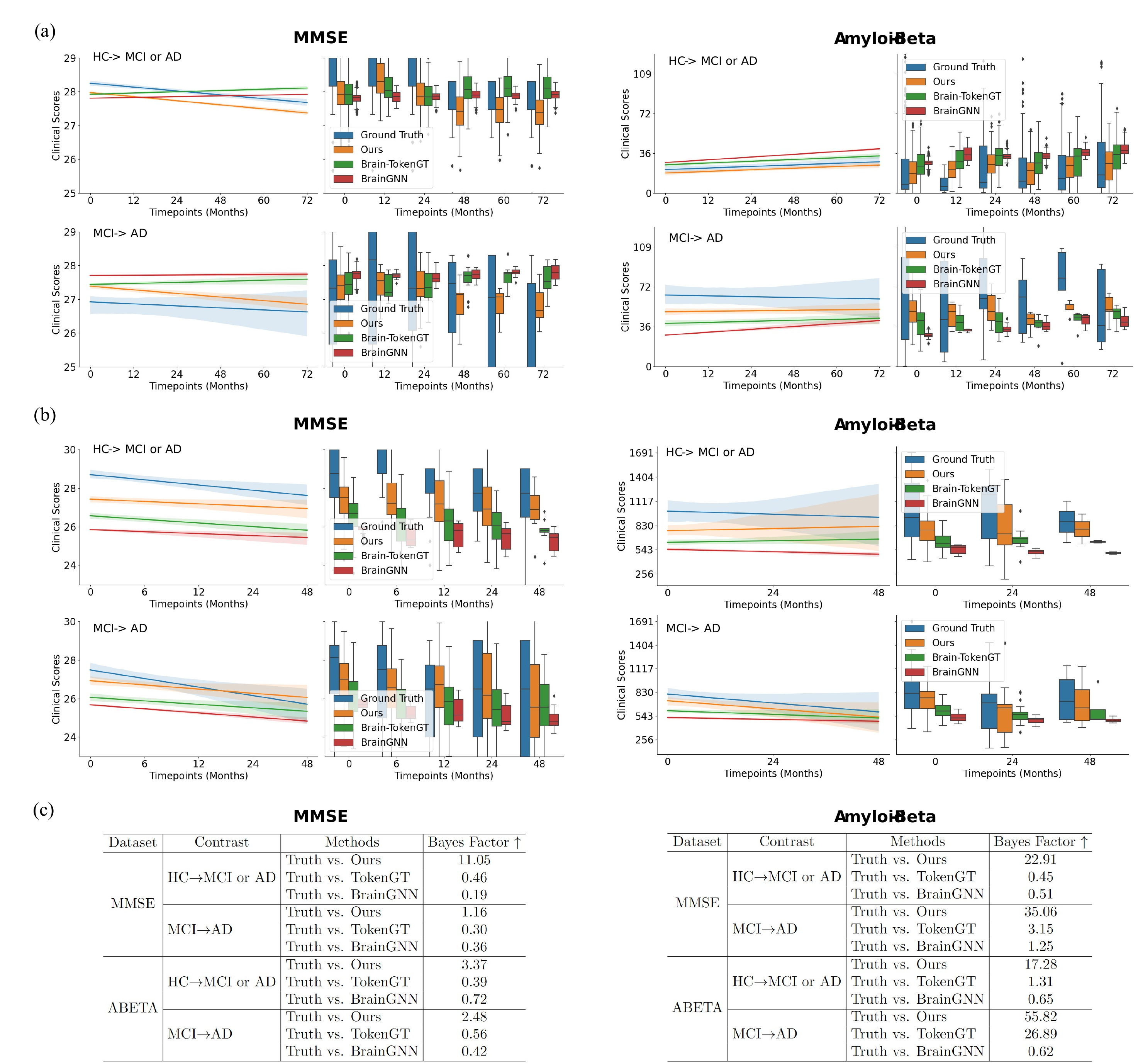}
\caption{(a) Prediction of MMSE and ABETA clinical scores in OASIS-3 cohort using linear mixed model based on the salient ROI feature representation learned by our method. The prediction performance was evaluated using 5-fold cross-validation. (b) Prediction of MMSE and ABETA clinical scores in validation ADNI cohort using the same strategy as OASIS. (c) Bayes Factor to compare the predicted clinical scores and ground truth between the state-of-the-art machine learning models and our proposed method under different contrasts and datasets. }
\label{fig:regression}
\end{figure*}

\subsection*{Associations between circuit abnormalities and longitudinal AD-related measures}
\label{sec:result_score}
To further investigate the clinical relevance of the identified circuit abnormalities, we examined their relationship with longitudinal AD-related clinical measures, including Mini-Mental State Examination (MMSE) scores and amyloid-beta (A$\beta$) levels. Using the learned salient topological patterns, we predicted these clinical measures in progressive subjects to assess the functional significance of our identified biomarkers. We first z-score normalized the learned topological features and used them to train a linear mixed-effects regression model, which was evaluated using 5-fold cross-validation. Fig. \ref{fig:regression}(a) presents the predicted versus actual MMSE and A$\beta$ scores across two progression stages in the OASIS-3 cohort. The actual A$\beta$ was calculated by measuring centiloid-scaled standardized uptake value ratios (SUVRs) across relevant cortical regions, with higher values indicating increased A$\beta$ accumulation and thus more severe Alzheimer's disease pathology. Fig. \ref{fig:regression}(b) shows the corresponding results for the ADNI cohort. The actual A$\beta$ represents the concentration of amyloid plaque burden in cerebrospinal fluid (CSF), where smaller values are associated with more severe Alzheimer's disease. The visualizations illustrate the predictive accuracy of our method for two disease transitions: HC to MCI/AD progression and MCI to AD progression. The fit of the regression line and the box plot for longitudinal data further indicate a strong correlation between the ground truth and predicted scores of our method. To statistically validate the significance of our predictions, we conducted Bayes Factor analysis for the correlation. Fig. \ref{fig:regression}(c) summarizes the numerical performance comparison between our method and other brain network-based approaches across all progressive subjects. Our approach consistently achieved the highest Bayes Factors across different clinical contrasts and cohorts, reinforcing the robustness of our biomarker-driven predictions. Specifically, in the OASIS-3 cohort, the Bayes Factor of our method for the HC to MCI/AD transition was 11.05 for MMSE and 3.37 for A$\beta$, while in the ADNI cohort, it was 22.91 for MMSE and 17.28 for A$\beta$. Notably, Bayes Factors for the MCI to AD transition in OASIS-3 cohort were generally lower, likely due to fewer progressive subjects and shorter available longitudinal time points. However, the identified circuit biomarkers demonstrated significant predictive performance across all clinical measures, underscoring their strong association with progressive MCI and AD symptoms. These results highlight the potential of our method for tracking neurodegenerative decline and support its applicability for early-stage AD prognosis.

\begin{figure*}[!t]
\centering
\includegraphics[width=\linewidth]{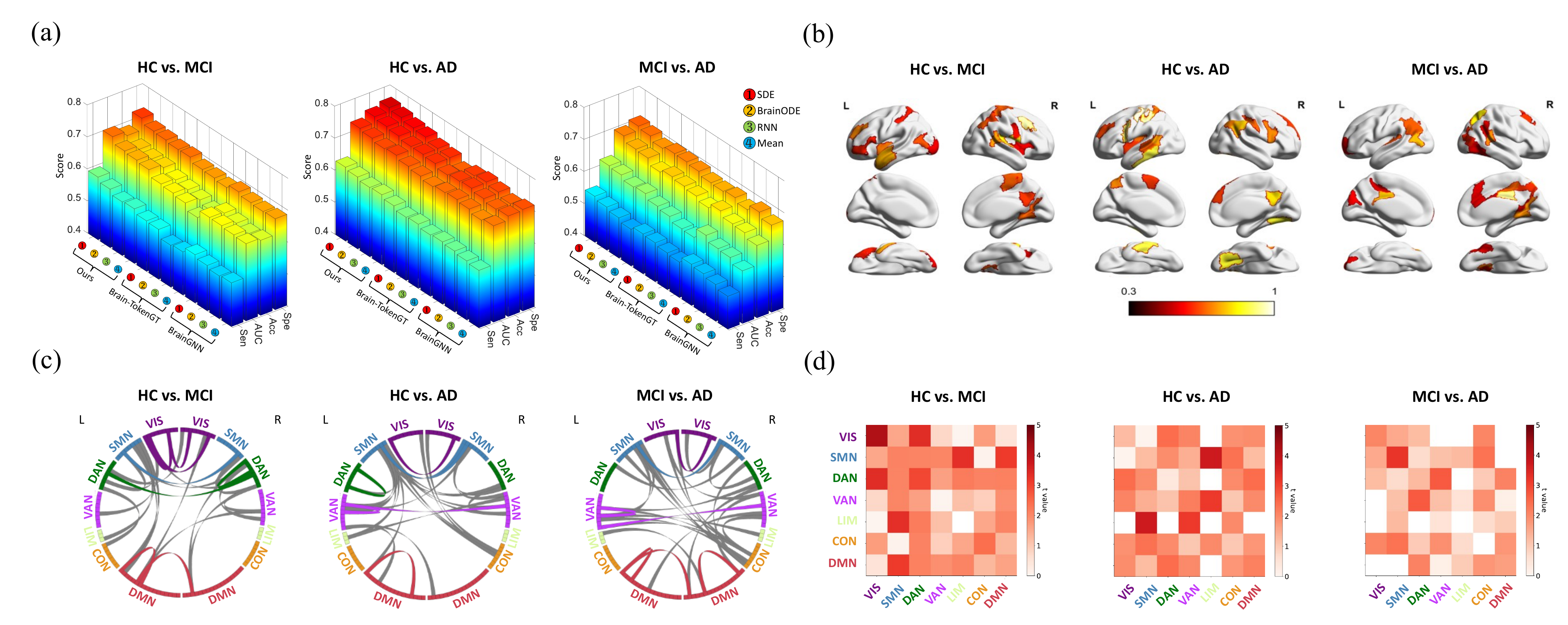}
\caption{{\bf Prediction and interpretation analysis to distinguish HC, MCI, and AD subjects in OASIS-3 cohort.} {\bf (a)} Classification comparison of three contrasts including HC vs. MCI, HC vs. AD, and MCI vs. AD between the state-of-the-art machine learning models and ours by using different preprocessing strategies including SDE, BrainODE, RNN, and Mean. The average classification accuracy (ACC), the area under the receiver operating characteristic curve (AUC), specificity (Spe), and sensitivity (Sen) were reported under the 5-fold cross validation.  {\bf (b)} Interpreting top 20 selected salient ROIs under three contrasts. The color bar ranges from 0.3 to 1.0. The bright-yellow color indicates a high score, while dark-red color indicates a low score. {\bf (c)} The top 30 most discriminative ROI connections are visualized for interpretation under three contrasts. The intra-network connections are colored based on the module itself and inter-network connections are colored in grey. {\bf (d)} Neural system-level interpretation of the most discriminative connections was reported under three contrasts. The dark-red color indicates significant connections. The non-significant connections are marked as white.}
\label{fig:diagnosis_label}
\end{figure*}

\subsection*{Longitudinal connectome-based prediction for AD diagnosis} 
We further evaluated the predictive capability of our method for AD diagnosis using all available longitudinal fMRI data. Fig. \ref{fig:diagnosis_label}(a) presents the classification performance across three contrasts: HC vs. MCI, HC vs. AD, and MCI vs. AD, comparing our approach with state-of-the-art machine learning models under different preprocessing strategies including SDE, BrainODE, RNN, and Mean. We report the average classification accuracy (ACC), area under the receiver operating characteristic curve (AUC), specificity (Spe), and sensitivity (Sen) using 5-fold cross-validation. Our method combined with the SDE preprocessing strategy achieved the best performance in distinguishing HC from AD/MCI, compared to baseline models. Fig. \ref{fig:diagnosis_label}(b) and \ref{fig:diagnosis_label}(c) provide interpretability analysis by highlighting the top salient ROIs and the most discriminative connections under the three classification contrasts. Additionally, Fig. \ref{fig:diagnosis_label}(d) reports neural system-level interpretations of the most discriminative connections. Specifically, we found that the extrastriate cortex in the visual network, and the parietal and post-central cortex in the dorsal attention network play a crucial role in the early stages of AD and MCI. The temporal lobe in the ventral attention network, and the cingulate and prefrontal cortex in the control network are essential for identifying severe cognitive impairment in AD. Additionally, auditory regions within the somatomotor network consistently emerge as salient regions across all three contrasts. For a detailed interpretation analysis from baseline to follow-up months under each contrast in the OASIS-3 cohort, refer to Supplementary Figs. \ref{fig:oasis_hcmci_interp}, \ref{fig:oasis_hcad_interp}, and \ref{fig:oasis_mciad_interp}.

\section*{Discussion}
Here we propose an advanced longitudinal connectome predictive modeling approach for uncovering critical neural circuits of AD progression. By incorporating a longitudinal spatio-temporal module, our approach enhances prediction accuracy and generalizability across both the OASIS-3 and ADNI cohorts, outperforming existing methods. Unlike traditional approaches that rely on either cross-sectional or regular longitudinal imaging data, our method leverages an SDE-based module to capture network dynamics, making it robust to irregularly-sampled longitudinal fMRI data and ensuring reliable performance across diverse scenarios and cohorts. Importantly, our method provides interpretable insights into AD progression by introducing sparse regional importance probabilities to identify salient ROIs and connective importance probabilities to highlight key brain network connections. These salient abnormal patterns are strongly associated with longitudinal AD-related clinical measures. Therefore, our method not only advances the predictability of AD progression, but also provides valuable interpretability for discovering neurological biomarkers and identifying functional brain connectivity abnormalities.

Our analysis revealed that the parahippocampal cortex, prefrontal cortex, and parietal lobule were consistently identified as salient ROIs across both the OASIS-3 and ADNI cohorts. Notably, these regions were more prominently detected when leveraging longitudinal data rather than baseline scans, suggesting that prediction accuracy improves as disease progression unfolds. The parahippocampal cortex, a region crucial for memory and recognition function, has been previously linked to functional changes in AD \cite{nigro2019functional} and has demonstrated greater dysfunction in AD patients compared to healthy controls \cite{chen2016differential}. Furthermore, parahippocampal atrophy has been recognized as an early biomarker for detecting AD/MCI \cite{mclachlan2018reduced}. The identification of the prefrontal cortex is consistent with studies associating protein alterations in this region with early cognitive impairment in AD \cite{moreno2020frontal, mahady2018frontal, montero2023proteomics}. Similarly, the parietal lobule, which exhibited high importance probabilities in our model, aligns with prior findings linking white matter hyperintensities in this region to an increased risk of AD \cite{brickman2015reconsidering}. These results suggest that the parahippocampal cortex, prefrontal cortex, and parietal lobule serve as early-stage biomarkers of AD/MCI progression. Additionally, other regions such as the extrastriate superior cortex, postcentral cortex, and insula also demonstrated high importance, further supporting their potential role in distinguishing progressive AD/MCI subjects.

Our analysis of functional brain connectivity abnormalities revealed that the most discriminative connections in AD progression were concentrated within the ventral attention network, dorsal attention network, and default mode network. These findings are consistent with prior studies reporting functional connectivity disruptions in the dorsal attention network during AD progression \cite{wu2022activation}. Furthermore, our results identified both intra- and inter-network abnormalities within the ventral and dorsal attention systems, aligning with fMRI studies that link functional degeneration in these networks to amnestic MCI and AD \cite{zhang2015functional}. Importantly, our findings reinforce the role of the default mode network (DMN) in AD progression, particularly through disruptions in the parahippocampal and prefrontal cortex, regions previously implicated in progressive cognitive impairment. The DMN has been shown to be among the earliest affected networks in autosomal dominant AD \cite{chhatwal2013impaired, malotaux2023default}, further supporting our observation that connectivity dysfunctions within this key network are strongly associated with cognitive decline. These results suggest that network-level abnormalities within the ventral attention, dorsal attention, and default mode networks play a critical role in the onset and progression of AD.

\begin{figure*}[!t]
\centering
\includegraphics[width=\linewidth]{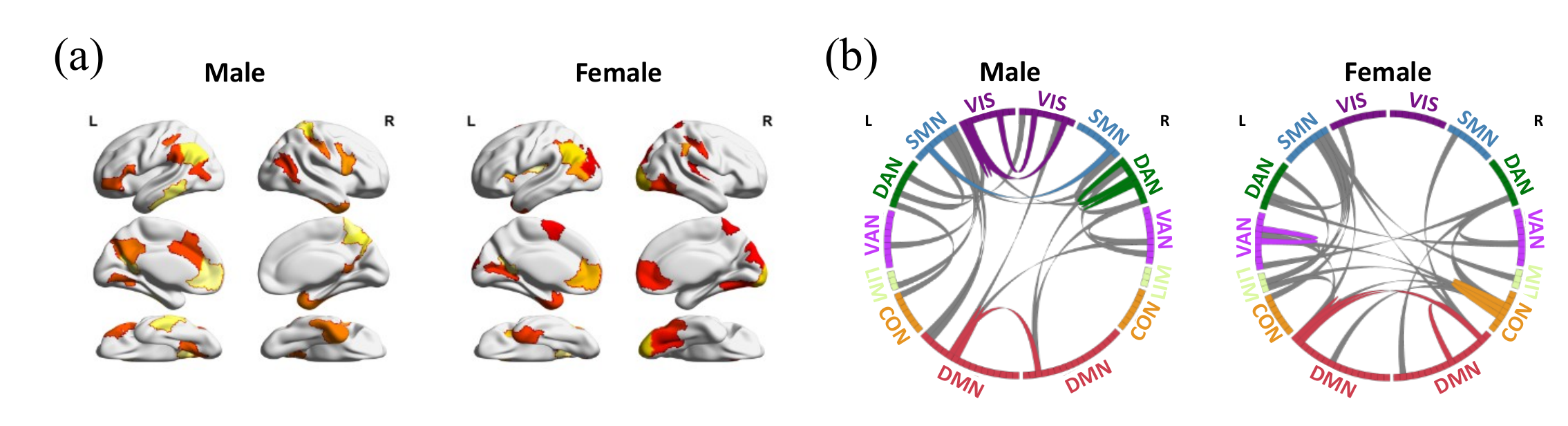}
\caption{\textbf{Sex-related differences}. Interpreting (a) top 20 salient ROIs and (b) the most discriminated connections in males and females. VIS = Visual Network, SMN = Somatomotor Network, DAN = Dorsal Attention Network, VAN = Ventral Attention Network, LIM = Limbic Network, CON = Control Network, DMN = Default Mode Network.}
\label{fig:sexrelated}
\end{figure*}

Our study further revealed sex-related differences in AD progression, demonstrating that females exhibit distinct biomarkers compared to males. Specifically, the temporal-parietal pole and parahippocampus were identified as the most salient ROIs in females (Fig. \ref{fig:sexrelated}), consistent with previous reports that females with AD experience greater cortical thinning in temporal regions \cite{cieri2022relationship}. the prefrontal cortex and cingulate cortex showed higher importance in females than in males, suggesting a strong association with MCI progression. The posterior cingulate cortex, in particular, has been recognized as a key region affected during the prodromal stage of AD \cite{scheff2015synaptic}. Our results suggest that cingulate atrophy in females may play a critical role in early dementia development and its progression to AD. While prior studies have linked these brain regions to cognitive impairment \cite{schwab2020functional, yu2017directed}, our study is the first to report these findings in a sex-specific longitudinal analysis of AD progression. This provides novel evidence on sex-related differences in biomarkers and their connectivity dysfunctions related to AD. For a detailed sex-specific interpretation from the baseline to follow-up timepoints, refer to Supplementary Figs. \ref{fig:oasis_female_interp} and \ref{fig:oasis_male_interp} for the OASIS-3 cohort, and \ref{fig:adni_female_interp} and \ref{fig:adni_male_interp} for the ADNI cohort.

In summary, we described a method to learn spatio-temporal network dysfunctions for predicting AD progression, even using irregular-sampled longitudinal fMRI data. Our interpretation result in two independent cohorts found that the parahippocampal cortex, prefrontal cortex, and parietal lobule were the potential biomarkers for AD progression. We further demonstrated that the prominent brain connectivity abnormalities both intra- and inter- ventral attention, dorsal attention, and default mode network were most important for distinguishing progressive subjects from stable ones. These disease-related network-based patterns identified by our method demonstrated strong predictability for longitudinal AD-related clinical measures. These findings suggest that our proposed framework can be effectively applied to predict disease progression and identify potential biomarkers for various neurological disorders, even under irregularly-sampled longitudinal analysis.

\section*{Methods}

\begin{figure*}[!t]
\centering
\includegraphics[width=\linewidth]{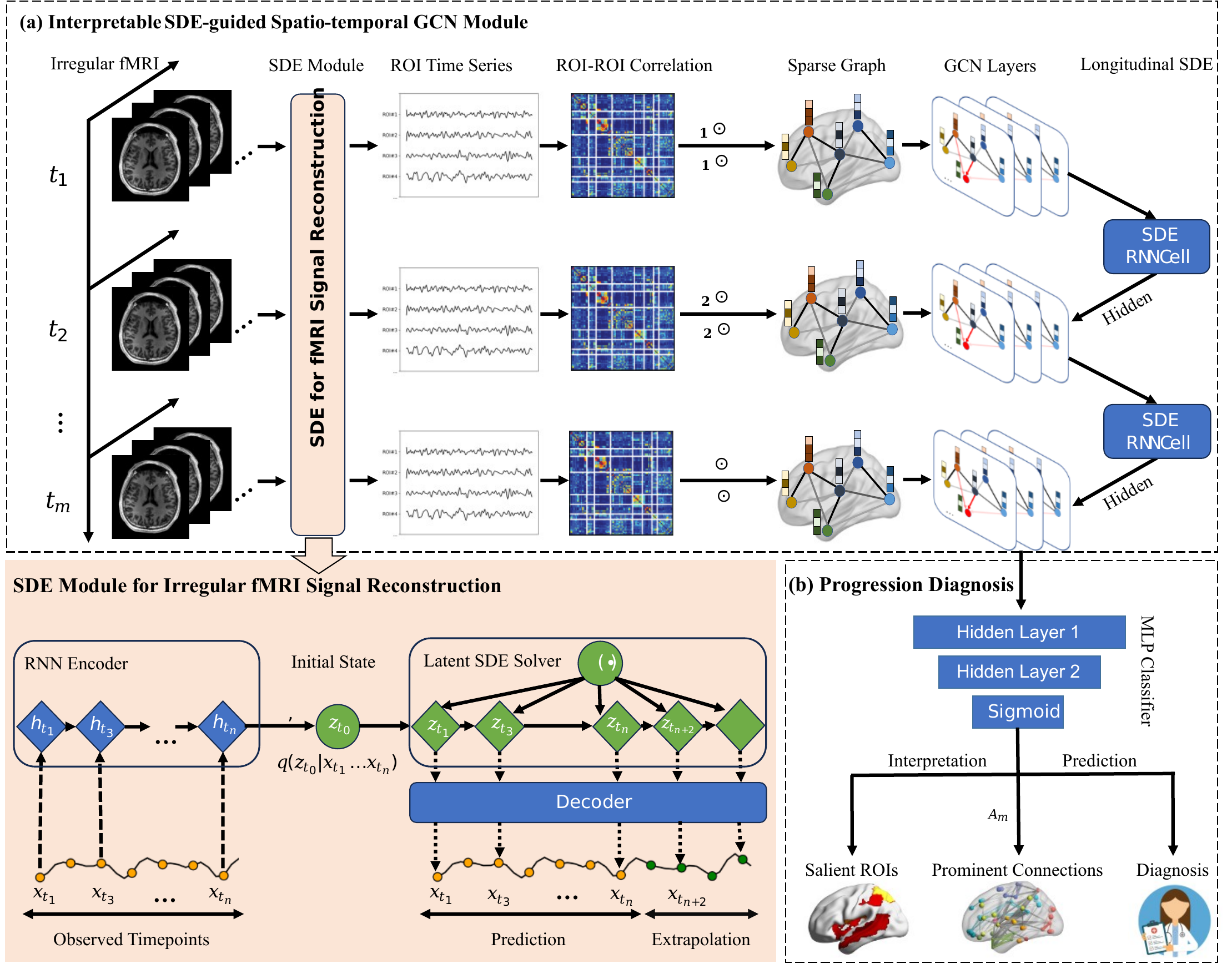}
\caption{A detailed pipeline of our proposed method for predicting Alzheimer's progression and discovering underlying biomarkers. {\bf a} The datasets consist of longitudinal fMRI scans, with time points ranging from the baseline ($t_1$) to the final time point ($t_m$, $m=6$). The longitudinal fMRI scans are preprocessed and reconstructed to continuous signals for each ROI by the SDE method. The dynamic graphs were built by using longitudinal reconstructed fMRI signals. First, the connections between ROIs were quantified by measuring the correlation between their reconstructed signals. Next, the reconstructed signals in the ROIs, along with their connections, were combined with the proposed learnable importance probabilities to generate a sparse graph. The dynamic graphs thus consist of longitudinal sparse graphs. The dynamic graphs are sent to our SDE-guided spatio-temporal GNN model to learn and evolve the longitudinal representation. {\bf b} The learned representation is fed into an MLP classifier to predict the disease progression. The importance probabilities on nodes and edges of sparse graphs provide the interpretation for the salient ROIs and the prominent disease-specific connections.}
\label{fig:architecture_overview_detailed}
\end{figure*}

\subsection*{Data preprocessing}
\label{method:data_acquisition}
In this work, we evaluated our framework on two longitudinal fMRI datasets including the Open Access Series of Imaging Studies (OASIS-3) \cite{marcus2010open} and the public Alzheimer's Disease Neuroimaging Initiative (ADNI) \cite{mueller2005alzheimer}. The OASIS-3 cohort consisted of 554 stable subjects and 193 progressive subjects. The ADNI cohort consisted of 261 stable subjects and 33 progressive subjects. Both cohorts include subjects with up to six time points. We conducted the prediction experiment in the contrast of stable vs. progressive subjects. Importantly, we only used time points before progression occurred, ensuring that our task focused on predicting future progressive subjects. See the Supplementary Tables \ref{table:demographics_oasis} and \ref{table:demographics_adni} for detailed demographics about age, gender, and clinical scores in both datasets.

The fMRI scan was preprocessed using fMRIPrep pipeline \cite{esteban2019fmriprep}. Preprocessing began with T1-weighted (T1w) image corrections, including intensity non-uniformity correction and skull stripping. The T1w reference was then spatially normalized through nonlinear registration \cite{avants2008symmetric}. Using FSL, brain tissues—cerebrospinal fluid, white matter, and gray matter were segmented from the skull-stripped T1w image \cite{zhang2000hidden}. Fieldmap data were employed to correct distortions in both low- and high-frequency components, producing a distortion-corrected echo-planar imaging (EPI) reference. This EPI reference was then co-registered with the anatomical T1w reference to enhance alignment accuracy. Subsequently, the blood-oxygenation-level-dependent (BOLD) reference was aligned to the T1w image using a boundary-based registration method with nine degrees of freedom, ensuring further correction of residual distortions \cite{greve2009accurate}. Head-motion correction was performed using MCFLIRT (FSL), which estimated rotation and translation parameters from volume-to-reference transformations. BOLD signals were then slice-time corrected, realigned to the participant’s native space while incorporating head-motion and susceptibility distortion corrections, and subsequently resampled into standard MNI152NLin2009cAsym space. To further refine the data, automatic removal of motion artifacts was applied using independent component analysis (ICA-AROMA) \cite{pruim2015ica}. This step was conducted after removing non-steady-state volumes and applying spatial smoothing with an isotropic Gaussian kernel of 6 mm full-width at half-maximum (FWHM). The final output was a fully preprocessed BOLD time series in standard MNI space, optimized for subsequent functional connectivity analyses.


\subsection*{Reconstructing the irregular fMRI signal}
We introduce an encoder-decoder based method with an SDE module to reconstruct the missing values in irregular dynamic fMRI signals. By capturing the trends and continuity within the temporal evolution of dynamic data, the SDE-based approach enables learning from irregular observations and facilitates continuous inference based on the learned dynamics. Therefore, we use the SDE module to infer a continuous latent representation of missing values in the signals. Specifically, we use an RNN as an encoder to learn the observed $n$ times from $x_1$ to $x_n$ to compute the approximate posterior $q$. The neural network $f$ is applied to transform the representation of RNN to the mean $\mu$ and standard deviation $\sigma$ of the initial state $z_0$.

\begin{equation}
q(z_o|x_1, x_2, ...,x_n)=\mathcal{N}(\mu, \sigma), \text{where }\mu, \sigma=f(RNN(x_1, x_2, ..., x_n))
\end{equation}

Given the initial state $z_0$ sampled from the approximate posterior $q$ learned by RNN encoder, we use a neural SDE model to generate the latent state at every timepoint.

\begin{equation}
\begin{split}
&z_0 \sim p(z_0) = q(z_o|x_1, x_2..., x_n)\\
&z_0, ..., z_t = SDESolver(g, z_0, (t_1, ..., t_n))
\end{split}
\end{equation}

where $t$ is the index of time, $n$ is the number of observations and $g$ is the neural SDE function. Finally, a decoder is applied to reconstruct the dynamic fMRI signals from the latent space $z$. Therefore, we build the encoder, generative model, and decoder together to model the trajectories of fMRI signals and generate the missing values. We use the mean squared error (MSE) loss to train the model by reconstructing the observed times, and add a Kullback–Leibler divergence term to regularize the latent initial states to meet the normal distribution and stabilize the training process.

\subsection*{Brain graph construction}
\label{method:construct_graph}
After reconstructing the irregular fMRI signals, we follow the standard to calculate the correlation coefficient of fMRI between ROIs to construct the brain connectivity \cite{li2021braingnn}. Due to the over-smoothing effect of the general graph neural networks for densely connected graphs \cite{cai2020note}, it is better to avoid dense graphs and thresholding correlation tends to lead to sparse graphs. Edges are defined by thresholding correlations to achieve sparse connections. In practice, we follow the previous study \cite{li2021braingnn} to use the top 10\% positive values which guarantees no isolated nodes in the graph. Therefore, we use the reconstructed fMRI signal as node features and correlations as the adjacency matrix in the brain connectivity graph. We construct the brain connectivity graph for all of the subjects. 


\subsection*{SDE-guided Spatial-temporal GNN model}
We introduced a spatial-temporal GNN model guided by SDE to obtain node embeddings that capture the spatial and temporal information of the functional connectome trajectory. With informative node features, we learned the dynamic graphs by treating parameters in graph convolutional layers as hidden states $\bm{H_{t_i}}$ at time $t_i$ of the dynamic system and used an RNN cell to update the hidden states $\bm{H_{t_i}}$. To learn the irregular-sampled longitudinal fMRI data, we add an SDESolver before RNN cell to update the hidden states $\bm{H_{t_i}}$ based on the time interval $\Delta t$ between irregular timepoints, where we define time interval $\Delta t = t_i - t_{i-1}$. Our architecture is composed of three types of layers: a neural SDE layer to learn the irregularity of timepoints, an RNN cell layer to learn the dynamic variation, and a graph convolutional layer to learn the graph topology. Combining them together, our method can effectively capture information from irregular-sampled dynamic graphs. Mathematically, our architecture can be expressed as:
\begin{equation}
\begin{split}
&\bm{H_{t_i}}^{'}=\bm{H_{t_{i-1}}}^{} +  \int _0^{\Delta t}f(H_\theta)\text{d}\theta={SDESolver}(f, \bm{H_{t_{i-1}}}^{}, \Delta t),\\ 
&\bm{H_{t_i}}=\text{RNNCell}(\bm{Z_{t_i}}, \bm{H_{t_i}}^{'})\\
&\bm{Z_{t_i}}=\text{GNN}(\bm{A_{t_i}}^{}, \bm{Z_{t_i}}, \bm{H_{t_i}})
\end{split}
\end{equation}

where $f$ is the neural network in the SDE function, $\bm{Z_{t_i}}$ is the learned node representation of the brain connectivity graph at time $t_i$, and $\bm{A_{t_i}}$ is the adjacency matrix. In graph topological learning, the Graph Neural Network can embed node-level features into a low dimensional space, and summarize them into graph-level features \cite{kipf2016semi}.  To learn the graph topology, the graph convolutional layer recursively updates a node representation by transforming and aggregating the neighboring feature vectors. Mathematically, the propagation update of node representation in our model can be calculated as:

\begin{equation}
\bm{Z_{t_i}}=\sigma(\bm{\widetilde{D}}^{-\frac{1}{2}} \bm{\widetilde{A}} \bm{\widetilde{D}}^{-\frac{1}{2}}\bm{Z_{t_i}}\bm{H_{t_i}^T})
\end{equation}

where $\bm{Z_0}=\bm{X}$, $\bm{Z_{t_i}} \in \mathbb{R}^{N \times d_{l}}$ is the output of the graph convolution layer, $N$ is the number of node, and $d_l$ is the number of  output channels. We add self-loops into the adjacency matrix $\bm{\widetilde{A}} = \bm{A}+\bm{I}$, where $\bm{I}\in \mathbb{R}^{N \times N}$ is the identity matrix. In this equation, $\bm{H_{t_i}}\in \mathbb{R}^{N \times d_{l}}$ are the learnable hidden states in dynamic graphs, $\bm{\widetilde{D}}$ is the diagonal degree matrix with  $\bm{\widetilde{D}}_{i,i} = \sum_{j}\widetilde{A}_{i,j}$, and $\sigma$ is the sigmoid function. Meanwhile, $\bm{\widetilde{A}}$ is normalized by multiplying $\bm{\widetilde{D}}^{-\frac{1}{2}}$, which can keep a fixed feature scale after graph convolution layer.

After the graph convolution layer, the node pooling layer is applied to group the node-level features together to summarize the graph-level features by using the global max pooling and global mean pool strategies. Next, the summarized output $\bm{Z_{t_i}}$ of the graph convolution layer is flattened into a single feature vector, which is fed into an MLP classifier with a sigmoid function for the final classification. Finally, we apply the supervised cross-entropy loss function for disease progression prediction.

\subsection*{Importance probabilities as the interpretation} 
We utilized the importance probabilities from the sparse interpretable GNN method \cite{zhou2022sparse} to explain GNN predictions, identify salient ROIs, and highlight the most discriminative connections related to AD progression. Given that the adjacency matrix $\bm{A}$ and regional feature $\bm{X}$ of original brain connectivity graph $G$ may contain redundant or noisy information, we introduce sparsity on them by learning a shared feature importance probability $\bm{P_X}$, and the individual edge importance probability $\bm{P_A}$ between nodes for each subject. Therefore, we mathematically expressed the final prediction output $\hat{y}$ of the GNN model $\Phi$ as: $\hat{y}=\Phi (\bm{A} \odot \bm{P_A}, \bm{X} \odot \bm{P_X})$. Generally, the problem of exploring the important subgraph and the important subset of node feature is translated into the inference of importance probability $\bm{P_A}$ on edges and $\bm{P_X}$ on nodes. The importance probabilities are applied to the dynamic functional brain network and node feature across all subjects.

Given that the different signal features of ROIs contribute differently to the disease progression, we define the feature importance probability $\bm{P_X}\in\mathbb{R}^{N \times D}$, where $\bm{P_X}=[\bm{p_1,p_2,...,p_N}]$, and $\bm{p_i}\in\mathbb{R}^D$, $1\leq i \leq N$, indicates the feature importance probability for each ROI, and $D$ is the dimension of signal. Because the regional features are associated with the strength of their connections, we define the edge importance probability $\bm{P_A}\in\mathbb{R}^{N \times N}$ for node $i$ and node $j$ by considering the joint connection between node features $\bm{x_i}$ and $\bm{x_j}$:
\begin{equation}
P_{A_{i,j}}=\sigma(\bm{v}^T[\bm{x_i} \odot \bm{p_i} || \bm{x_j} \odot \bm{p_j}])
\end{equation}

where $\bm{v}\in\mathbb{R}^{2D}$ is the learnable parameter, $\bm{p_i}$ is feature importance probability of node $i$, $\odot$ denotes the Hadamard element-wise product function, and $||$ denotes the concatenation function. In this equation, the edge importance probability is calculated by combining node feature and their feature probabilities together. This mechanism is beneficial to identify the prominent brain connections based on the information of the node features.

\subsection*{Loss functions} 
In our loss functions, we used the cross-entropy loss for the classification task, and combined three regularization losses including the mutual information loss to determine the $\bm{P_X}$ and $\bm{P_A}$, as well as the ${\ell}_1$ and entropy regularization loss to promote the sparsity on them. The interpretation of the GNN model is generated by exploring the important subgraph $G_s$ and the important subset of node feature $X_s$, which have the maximum mutual information with the distribution of truth labels. Specifically, we train our model and find the importance probability $\bm{P_X}$ and $\bm{P_A}$ by maximizing the mutual information between the true label $y$ and the predictive output $\hat{y}$ learned from the subgraph $G_s$ and the subset of node feature $X_s$. We define that there are $C$ classes, and the mutual information loss $\mathcal{L}_{MI}$ is expressed as:

\begin{equation}
\mathcal{L}_{MI}=-\sum_{c=1}^{C}\mathbbm{1}[y=c]\log P_\Phi (\hat{y}=y\:|\:G_s=\bm{A} \odot \bm{P_A},X_s=\bm{X} \odot \bm{P_X})
\end{equation}

Our method minimized the mutual information loss to determine the importance probability $\bm{P_X}$ and $\bm{P_A}$ for progression prediction. The ${\ell}_1$ and entropy regularization losses were further applied in order to induce sparsity on $\bm{P_X}$ and $\bm{P_A}$. The sparsity loss $\mathcal{L}_{SP}$ on $\bm{P_X}$ and $\bm{P_A}$ can be expressed as:

\begin{equation}
\begin{split}
\mathcal{L}_{SP}= \left\| \bm{P_X} \right\|_1+\left\| \bm{P_A} \right\|_1
\end{split}
\end{equation}

Meanwhile, in order to encourage discreteness in the probability distribution, the element-wise entropy regularization loss is applied:

\begin{equation}
\begin{split}
&\mathcal{L}_{P_X}=-(\bm{P_X}\log (\bm{P_X})+(1-\bm{P_X})\log (1-\bm{P_X}))\\
&\mathcal{L}_{P_A}=-(\bm{P_A}\log (\bm{P_A})+(1-\bm{P_A})\log (1-\bm{P_A}))\\
&\mathcal{L}_{EN}=\mathcal{L}_{P_A}+\mathcal{L}_{P_X}\\
\end{split}
\end{equation}

where $\mathcal{L}_{P_X}$ denotes the entropy loss on $\bm{P_X}$, $\mathcal{L}_{P_A}$ denotes the entropy loss on $\bm{P_A}$, and $\mathcal{L}_{EN}$ is the sum of both entropy losses. Both ${\ell}_1$ and entropy regularization loss induce sparsity on $\bm{P_X}$ and $\bm{P_A}$, which drive the unimportant or noisy entries towards zero. In the meanwhile, the mechanism of entropy regularization loss induces important features and connections to have higher probabilities towards one for disease progression. 

After combining all of the loss functions, the final loss function $\mathcal{L}$ can be expressed as:

\begin{equation}
\mathcal{L}=\mathcal{L}_{CE}+\lambda_1 \mathcal{L}_{MI}+\lambda_2 \mathcal{L}_{SP}+\lambda_3 \mathcal{L}_{EN}
\end{equation}

Where $\mathcal{L}_{CE}$ is the cross-entropy loss for the classification task and $\lambda$'s are the tunable hyperparameters as the penalty coefficients for the different losses. The effect of different $\lambda$ in the loss terms has been analyzed in the sparse GNN method \cite{zhou2022sparse}.

We trained the model and optimized the loss function to determine learnable parameters $\bm{P_X}$ and $\bm{P_A}$ for predicting AD progression. In our result, the importance probability $\bm{P_X}$ and $\bm{P_A}$ learned from model provide the interpretation of the salient ROIs and the prominent disease-specific brain connectivity abnormalities.

\subsection*{Model training and evaluation protocols}
We trained and tested our proposed method on Pytorch framework by using a Nvidia RTX A5000 with 24GB GPU memory. In the fMRI reconstruction, we set the hidden dimension of RNN encoder to 32, and the hidden dimension of neural network $f$ in SDE to 16. In the dynamic graph learning, we built the SDE-guided spatial-temporal GNN followed by three fully-connected layers, a dropout layer, and a sigmoid function with parameters $N=100$, $d_l=16$, and $C=2$. The hidden dimensions of the three fully-connected layers are 64, 16, and 1 respectively. The dropout rate is 0.5. In the experiment, we use grid search to find the best hyperparameters. We trained our method for 100 epochs with a learning rate of 0.001. AdamW was used as the learning optimizer. Each batch contained 32 dynamic graphs during training. Meanwhile, we performed 5-fold cross-validation to examine performance and reported average classification accuracy, the area under the receiver operating characteristic curve (ROC-AUC), sensitivity, and specificity. 

\newpage

\bibliography{refs}



\section*{Acknowledgements}

This work was in part supported by NIH (R01LM013519, RF1AG077820, R21AG080425), NSF (IIS-2319451, MRI-2215789), DOE (DE-SC0025801), Lehigh University (CORE and RIG), and the Alzheimer's Association (AARG-22-972541).

\section*{Author contributions statement}


\section*{Declaration of Competing Interest}
The authors declare no competing financial interests.

\section*{Data and code availability}
All brain imaging datasets used in this study come from the public Alzheimer’s Disease Neuroimaging Initiative (ADNI) Website (https://adni.loni.usc.edu/). Codes of the sparse interpretable graph neural networks and brain connectome analyses will be deposited in an open-access platform, upon acceptance of this manuscript.

%



\newpage
\nolinenumbers
\section*{Supplementary Material}

\setcounter{table}{0}
\setcounter{figure}{0}
\renewcommand{\thetable}{S\arabic{table}}
\renewcommand{\thefigure}{S\arabic{figure}}


\begin{figure*}[!b]
\centering
\includegraphics[width=0.8\linewidth]{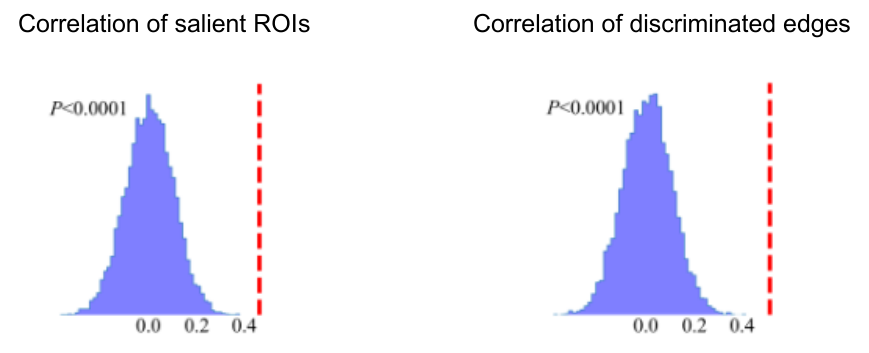}
\caption{The significant correlation of interpreted biomarkers between OASIS-3 and ADNI cohorts was confirmed by random permutation tests of 10,000 times. The actual correlation coefficient of interpreted biomarkers between the OASIS-3 and ADNI cohorts was 0.46 in salient ROIs and 0.55 in discriminated edges, which was indicated by red dashed lines.}
\label{fig:corr_biomarker}
\end{figure*}

\begin{table*}[!ht]
\renewcommand\arraystretch{1.3}
\setlength\tabcolsep{3pt}
  \centering
  \caption{ Demographics and the clinical scores for subjects in OASIS-3 cohort. The mean and standard deviation of Age, MMSE, and ABETA were reported in stable and progressive groups. Groups comparison in Gender was performed using Chi-square test. Groups comparison in Age, MMSE, and ABETA was performed using t-test. Their statistics and significant $p$ value were reported. F denotes female, and M denotes male.}
  \label{table:demographics_oasis}
  \scalebox{1}{
  \begin{tabular}{l|l|l|l|l}
  \hline
Category& Stable Group& Progressive Group& Statistics& p-value\\ \hline
  Sample Size&554&193& -&-\\ 
  Gender (M/F)&231/323 &106/87 & 2.31&2.47E-01\\
  Age (Mean $\pm$ std)&72.3 $\pm8.2$ &	74.5 $\pm9.1$& 3.10 & 1.32E-03\\
  MMSE (Mean $\pm$ std)&29.1 $\pm1.2$&27.2 $\pm2.5$& 17.37& 1.04E-61\\
  ABETA (Mean $\pm$ std)&25.6 $\pm31.2$&63.3 $\pm40.1$ & 11.20& 4.96E-27\\
  \hline
  \end{tabular}}
\end{table*} 

\begin{table*}[!ht]
\renewcommand\arraystretch{1.3}
\setlength\tabcolsep{3pt}
  \centering
  \caption{ Demographics and the clinical scores for subjects in ADNI cohort. The mean and standard deviation of Age, MMSE, and ABETA were reported in stable and progressive groups. Groups comparison in Gender was performed using Chi-square test. Groups comparison in Age, MMSE, and ABETA was performed using t-test. Their statistics and significant $p$ value were reported. F denotes female, and M denotes male.}
  \label{table:demographics_adni}
  \scalebox{1}{
  \begin{tabular}{l|l|l|l|l}
  \hline
Category& Stable Group& Progressive Group& Statistics& p-value\\ \hline
  Sample Size&261&33& -&-\\ 
  Gender (M/F)&118/143 &22/11 & 1.12&2.65E-01\\
  Age (Mean $\pm$ std)&71.5 $\pm7.1$ &	72.9 $\pm7.6$& 2.01 & 1.52E-02\\
  MMSE (Mean $\pm$ std)&27.7 $\pm2.8$&26.5 $\pm2.6$& 3.87& 1.19E-04\\
  ABETA (Mean $\pm$ std)&875.7 $\pm318.8$&725.2 $\pm325.2$ & 2.30& 2.24E-03\\
  \hline
  \end{tabular}}
\end{table*}


\begin{figure*}[!t]
\centering
\includegraphics[width=0.8\linewidth]{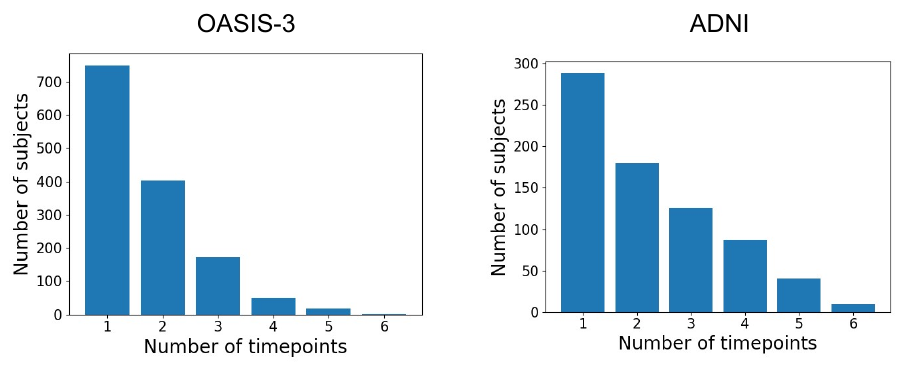}
\caption{Distribution of the number of timepoints for all subjects in the OASIS-3 and ADNI cohort. Note that during training, the model incorporates all preceding available timepoints rather than using only the current one.}
\label{fig:dist_of_timepoints}
\end{figure*}

\begin{table}[!ht]
\renewcommand\arraystretch{1.3}
\setlength\tabcolsep{3pt}
  \centering
  \caption{DeLong test to compare the area under the ROC curves (ROC-AUC) between the state-of-the-art machine learning models and our proposed method by using the same SDE reconstruction strategy and learning different numbers of longitudinal timepoints.}
  \label{table:ablation}
  \scalebox{1}{
  \begin{tabular}{c|l|cc|cc}
  \hline
\multirow{2}{*}{Timepoints}&\multirow{2}{*}{Comparing Methods}& \multicolumn{2}{c|}{OASIS} & \multicolumn{2}{c}{ADNI} \\\cline{3-6}
&&Z & $p$-value&Z & $p$-value\\\hline
   \multirow{2}{*}{1}
   &{Ours vs. BrainGNN}& 1.07 & 0.06& 2.89 & $<$0.01\\
   &{Ours vs. TokenGT}& 2.19 & 0.03& 1.72 & 0.05\\
   \hline %
   \multirow{2}{*}{2}
   &{Ours vs. BrainGNN}& 2.62 & $<$0.01& 2.11 & 0.03\\
   &{Ours vs. TokenGT}& 2.35 & 0.02& 2.74 & 0.01\\
   \hline %
   \multirow{2}{*}{3}
   &{Ours vs. BrainGNN}& 3.24 & $<$0.01& 3.02 & $<$0.01\\
   &{Ours vs. TokenGT}& 2.91 & $<$0.01& 2.57 & 0.01\\\hline %
   \multirow{2}{*}{4}
   &{Ours vs. BrainGNN}& 4.03 & $<$0.01& 3.36 & $<$0.01\\
   &{Ours vs. TokenGT}& 3.82 & $<$0.01& 3.79 & $<$0.01\\\hline
   \multirow{2}{*}{5}
   &{Ours vs. BrainGNN}& 2.93 & $<$0.01& 3.45 & $<$0.01\\
   &{Ours vs. TokenGT}& 3.26 & $<$0.01& 2.97 & $<$0.01\\\hline
   \multirow{2}{*}{6}
   &{Ours vs. BrainGNN}& 3.79 & $<$0.01& 3.03 & $<$0.01\\
   &{Ours vs. TokenGT}& 3.48 & $<$0.01& 3.68 & $<$0.01\\\hline
  \end{tabular}
  }
\end{table}


\newpage
\begin{figure*}[!t]
\centering
\includegraphics[width=\linewidth]{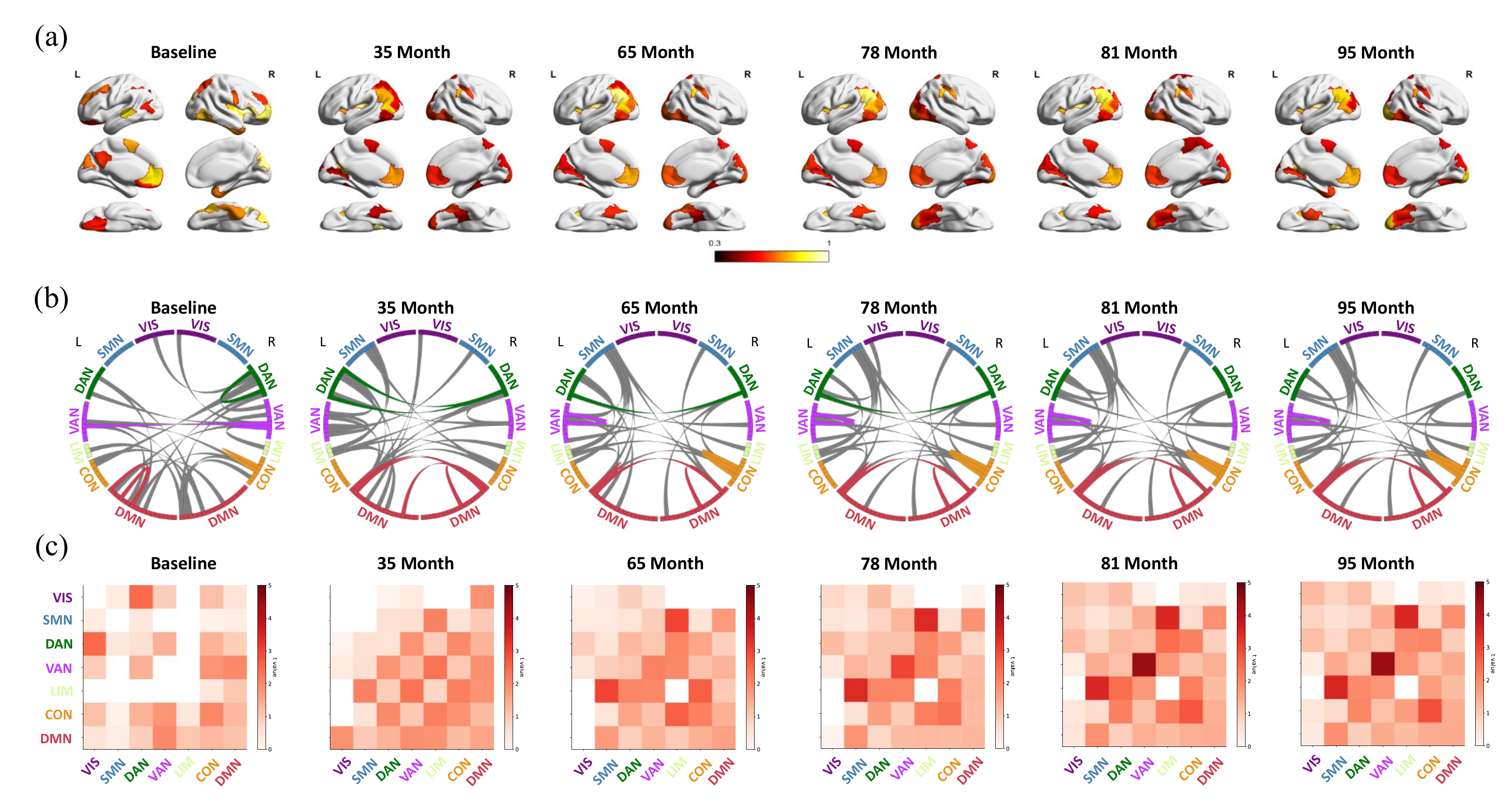}
\caption{{\bf Interpretation analysis of females in OASIS-3 cohort.} We analyzed the sex-related interpretation using the same interpretable strategy, reporting (a) salient ROIs, (b) the most discriminative connections, and (c) neural system-level dysfunctions. Our result indicated that the parahippocampus, prefrontal cortex, and cingulate cortex were the salient ROIs. The significant dysfunctions in AD progression were concentrated in dorsal attention network at the early stage, and in somatomotor, ventral attention, and default mode networks at the later stage. VIS = Visual Network, SMN = Somatomotor Network, DAN = Dorsal Attention Network, VAN = Ventral Attention Network, LIM = Limbic Network, CON = Control Network, DMN = Default Mode Network.}
\label{fig:oasis_female_interp}
\end{figure*}
\begin{figure*}[!t]
\centering
\includegraphics[width=\linewidth]{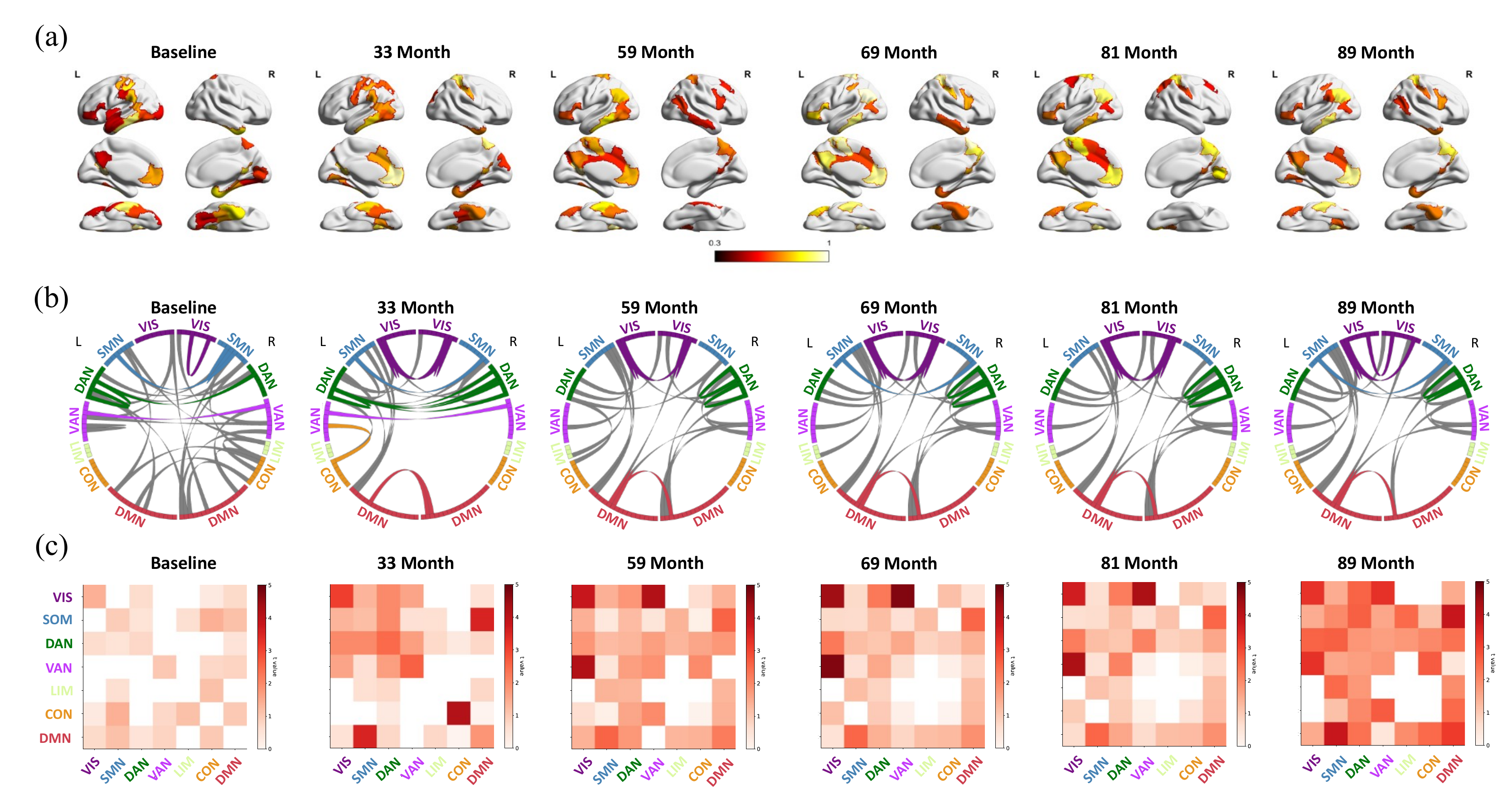}
\caption{{\bf Interpretation analysis of males in OASIS-3 cohort.} We analyzed the sex-related interpretation using the same interpretable strategy, reporting (a) salient ROIs, (b) the most discriminative connections, and (c) neural system-level dysfunctions. Our result indicated that the significant dysfunctions in AD progression were concentrated in visual, somatomotor, and ventral attention networks. VIS = Visual Network, SMN = Somatomotor Network, DAN = Dorsal Attention Network, VAN = Ventral Attention Network, LIM = Limbic Network, CON = Control Network, DMN = Default Mode Network.}
\label{fig:oasis_male_interp}
\end{figure*}
\begin{figure*}[!t]
\centering
\includegraphics[width=\linewidth]{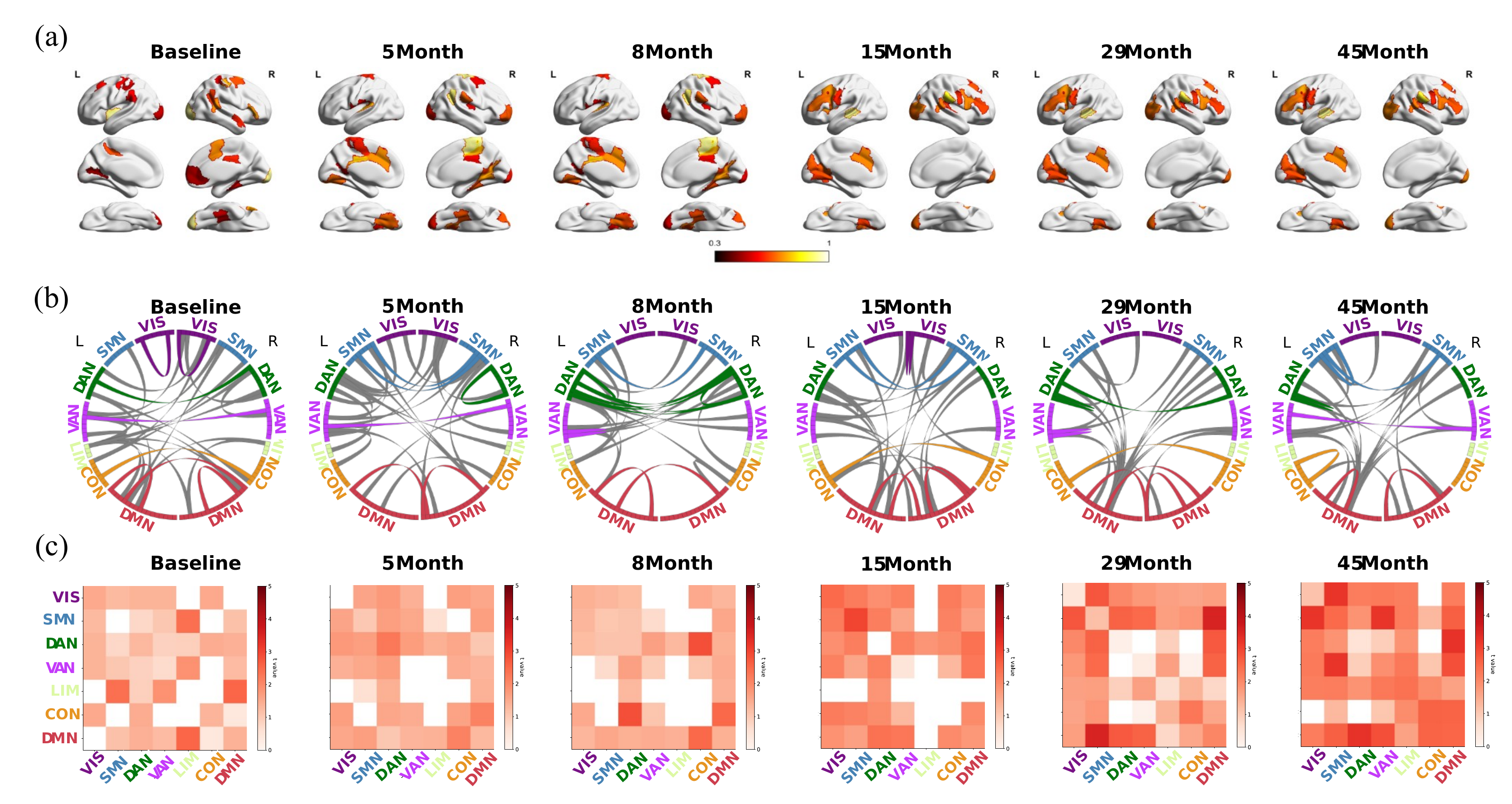}
\caption{{\bf Interpretation analysis of females in ADNI cohort.} We analyzed the sex-related interpretation using the same interpretable strategy, reporting (a) salient ROIs, (b) the most discriminative connections, and (c) neural system-level dysfunctions. Our result indicated that the significant dysfunctions in AD progression were concentrated in somatomotor, ventral attention, and default mode networks. VIS = Visual Network, SMN = Somatomotor Network, DAN = Dorsal Attention Network, VAN = Ventral Attention Network, LIM = Limbic Network, CON = Control Network, DMN = Default Mode Network.}
\label{fig:adni_female_interp}
\end{figure*}
\begin{figure*}[!t]
\centering
\includegraphics[width=\linewidth]{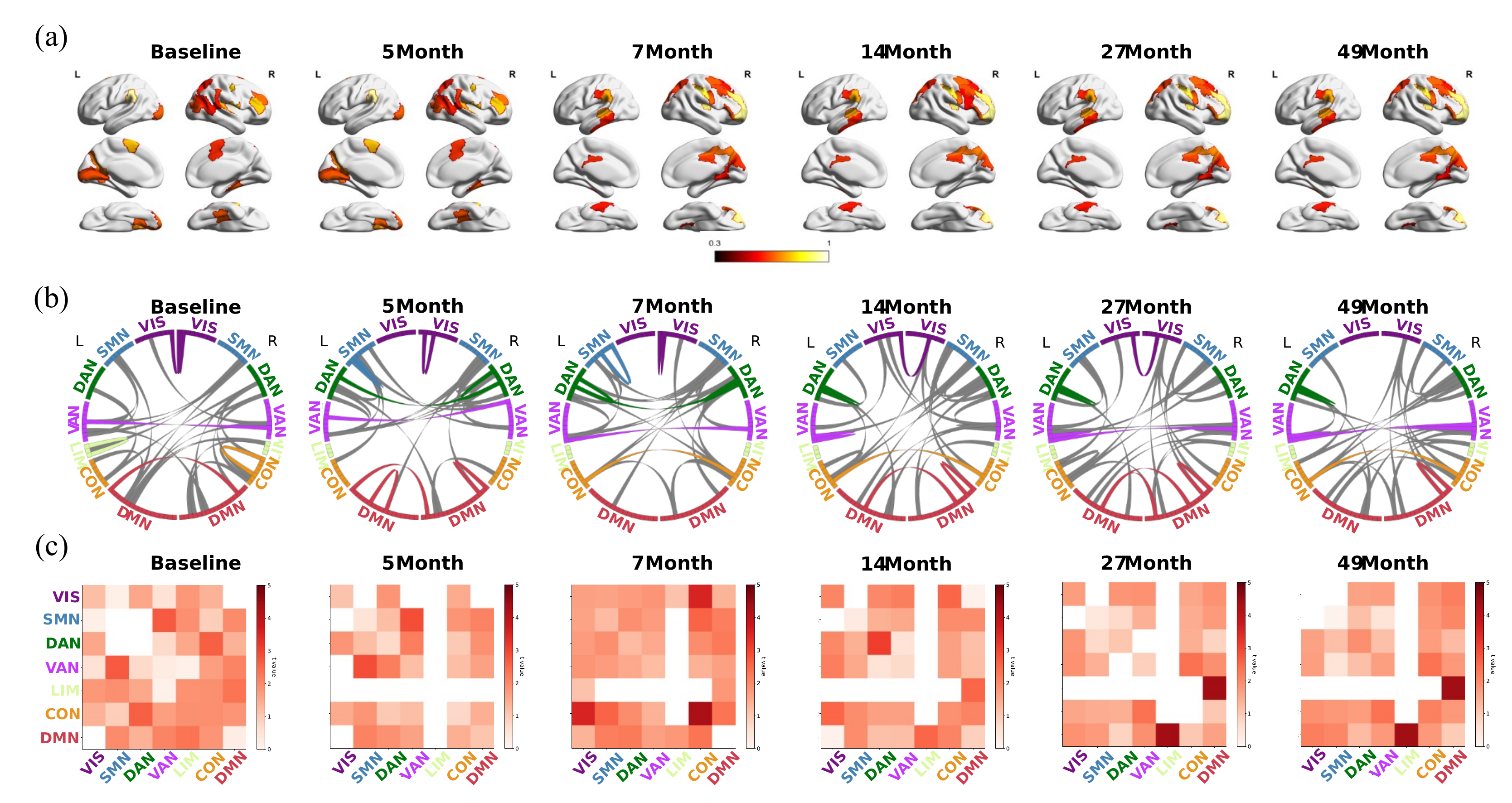}
\caption{{\bf Interpretation analysis of males in ADNI cohort.} We analyzed the sex-related interpretation using the same interpretable strategy, reporting (a) salient ROIs, (b) the most discriminative connections, and (c) neural system-level dysfunctions. Our result indicated that the significant dysfunctions in AD progression were concentrated in somatomotor, dorsal attention, and default mode networks. VIS = Visual Network, SMN = Somatomotor Network, DAN = Dorsal Attention Network, VAN = Ventral Attention Network, LIM = Limbic Network, CON = Control Network, DMN = Default Mode Network..}
\label{fig:adni_male_interp}
\end{figure*}
\begin{figure*}[!t]
\centering
\includegraphics[width=\linewidth]{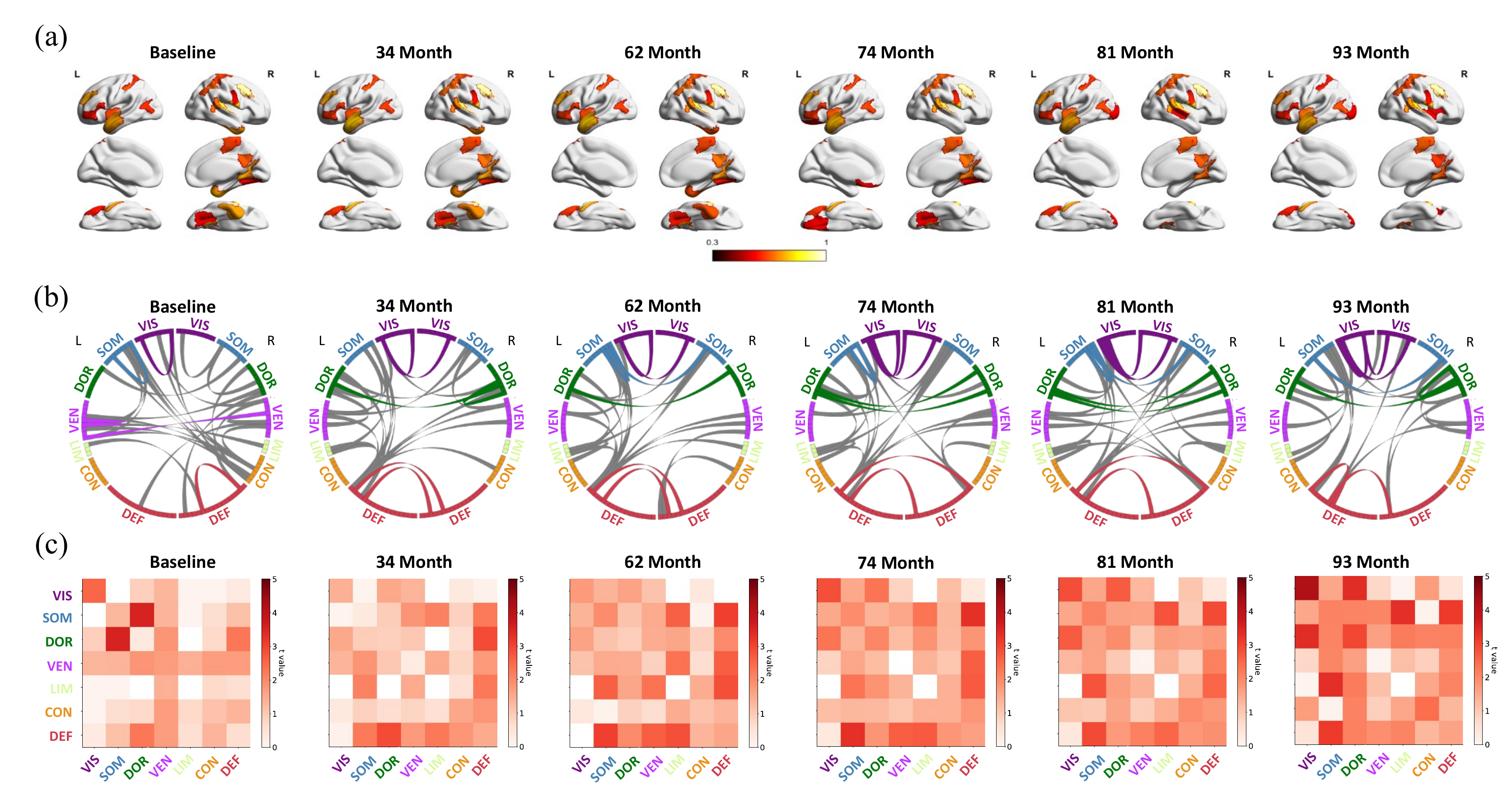}
\caption{{\bf Interpretation analysis of HC vs. MCI contrast in OASIS-3 cohort.} We analyzed the diagnostic interpretation using the same interpretable strategy, reporting (a) salient ROIs, (b) the most discriminative connections, and (c) neural system-level dysfunctions. Our result indicated that the significant dysfunctions in MCI were concentrated in visual, somatomotor, dorsal attention, and default mode networks. VIS = Visual Network, SMN = Somatomotor Network, DAN = Dorsal Attention Network, VAN = Ventral Attention Network, LIM = Limbic Network, CON = Control Network, DMN = Default Mode Network.}
\label{fig:oasis_hcmci_interp}
\end{figure*}
\begin{figure*}[!t]
\centering
\includegraphics[width=\linewidth]{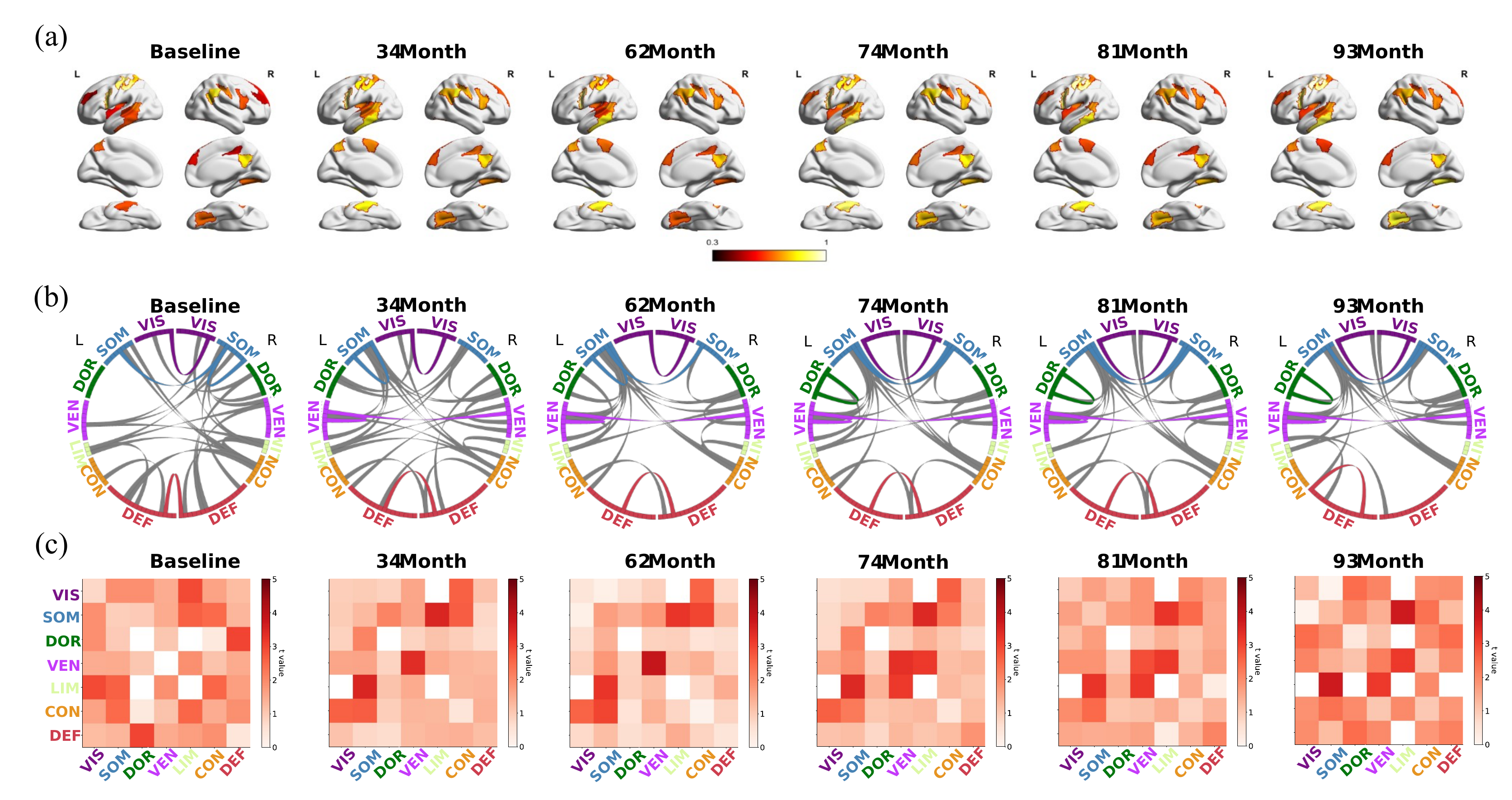}
\caption{{\bf Interpretation analysis of HC vs. AD contrast in OASIS-3 cohort.} We analyzed the diagnostic interpretation using the same interpretable strategy, reporting (a) salient ROIs, (b) the most discriminative connections, and (c) neural system-level dysfunctions. Our result indicated that the significant dysfunctions in AD were concentrated in somatomotor, and ventral attention networks. VIS = Visual Network, SMN = Somatomotor Network, DAN = Dorsal Attention Network, VAN = Ventral Attention Network, LIM = Limbic Network, CON = Control Network, DMN = Default Mode Network.}
\label{fig:oasis_hcad_interp}
\end{figure*}
\begin{figure*}[!t]
\centering
\includegraphics[width=\linewidth]{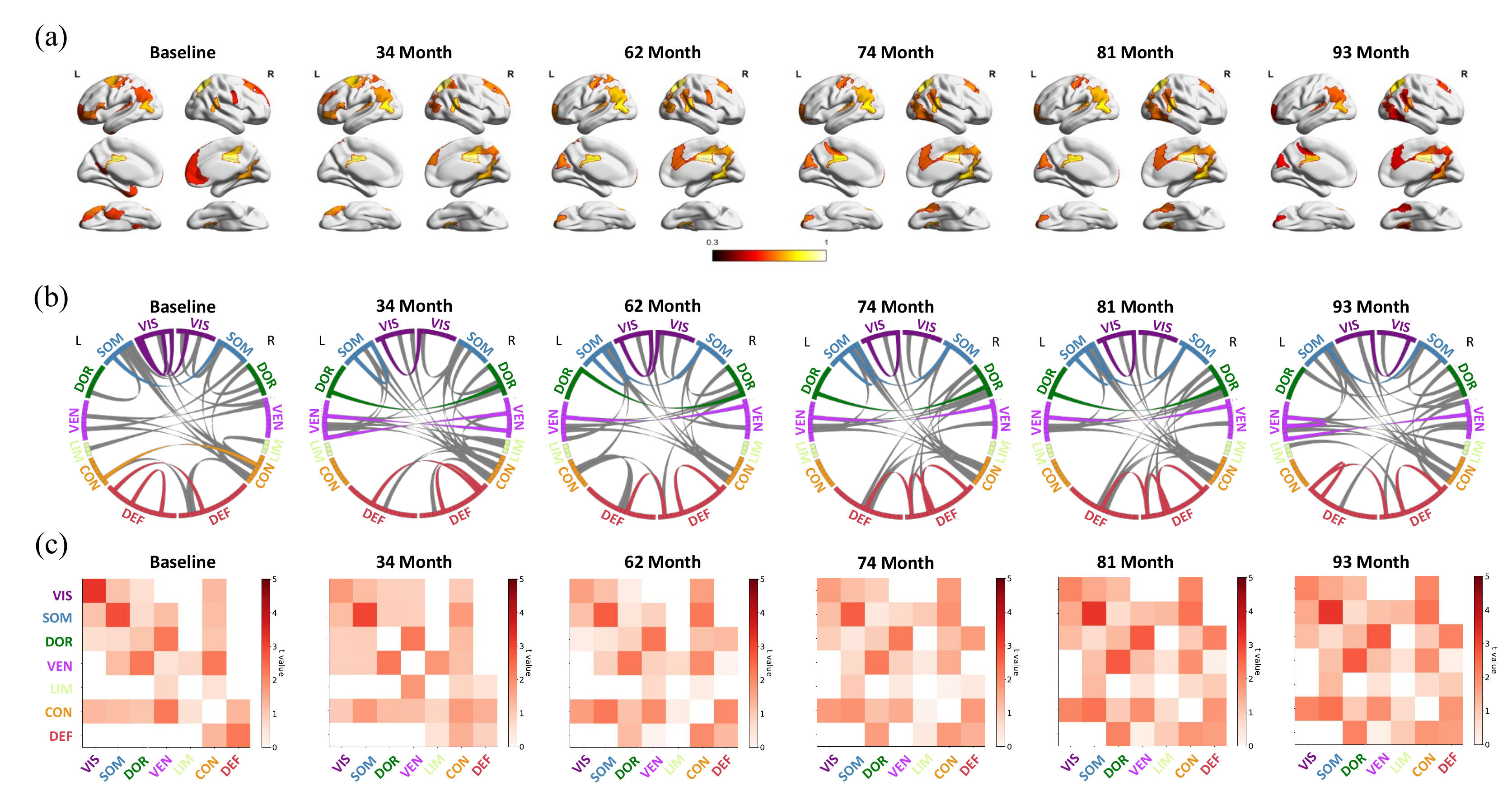}
\caption{{\bf Interpretation analysis of MCI vs. AD contrast in OASIS-3 cohort.} We analyzed the diagnostic interpretation using the same interpretable strategy, reporting (a) salient ROIs, (b) the most discriminative connections, and (c) neural system-level dysfunctions. Our result indicated that the significant dysfunctions in severe cognitive impairment were concentrated in somatomotor, dorsal attention, ventral attention, and default mode networks. VIS = Visual Network, SMN = Somatomotor Network, DAN = Dorsal Attention Network, VAN = Ventral Attention Network, LIM = Limbic Network, CON = Control Network, DMN = Default Mode Network.}
\label{fig:oasis_mciad_interp}
\end{figure*}
\begin{figure*}[!ht]
\centering
\includegraphics[width=\linewidth]{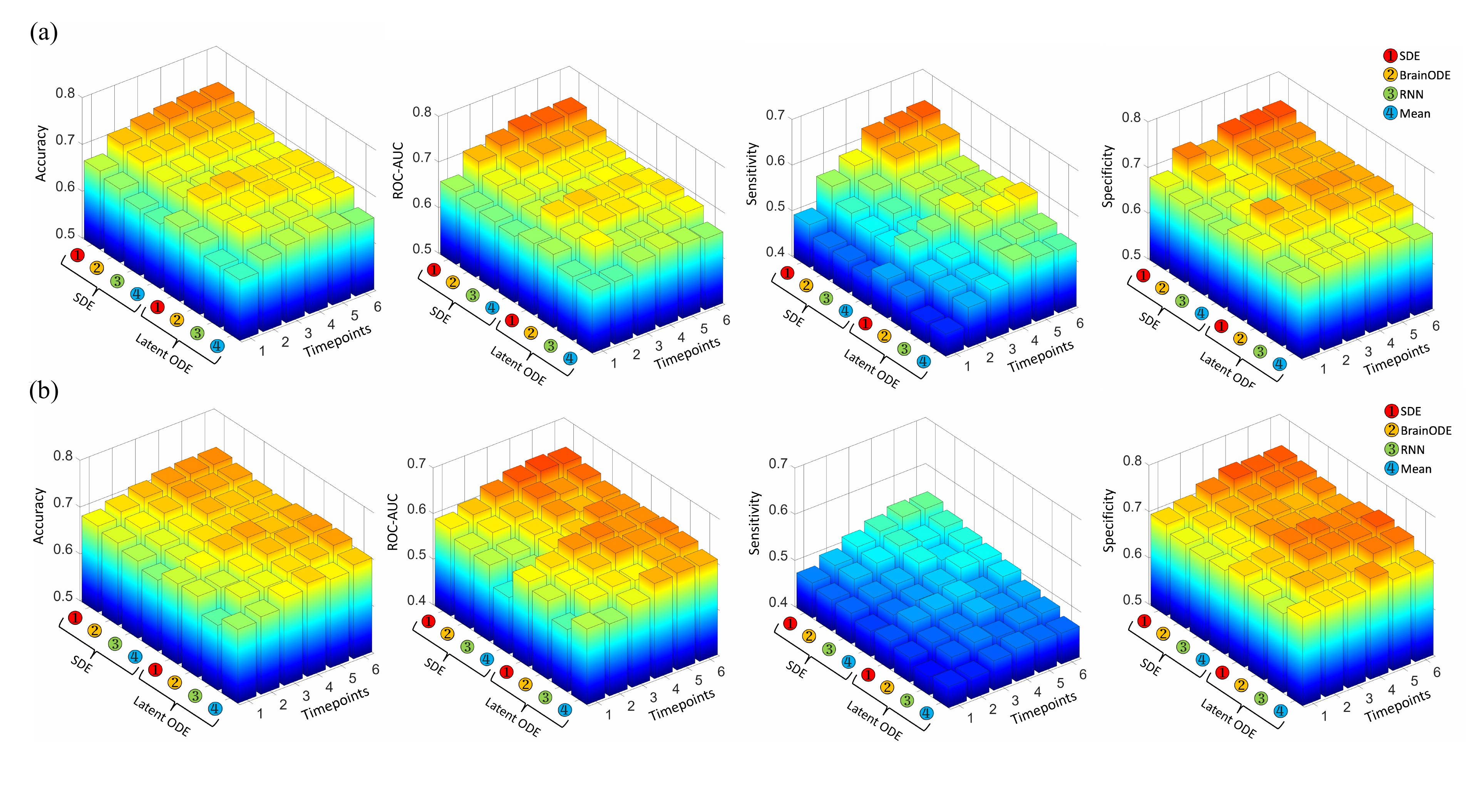}
\caption{Ablation study of spatial-temporal GNN module guided by SDE and latent ODE. The average classification accuracy, ROC-AUC, specificity and sensitivity are reported by using SDE and latent ODE module for both (a) OASIS and (b) ADNI dataset.}
\label{fig:ablation_study}
\end{figure*}

\end{document}